\documentclass[10pt,journal,compsoc]{IEEEtran}
\usepackage{amsmath}
\usepackage{amssymb}
\usepackage{array}
\usepackage{graphicx}
\usepackage[usenames, dvipsnames]{color}
\usepackage{rotating}
\usepackage{multirow}
\usepackage{pgfplots}
\usepackage{algorithm}
\usepackage{algorithmicx}
\usepackage[noend]{algpseudocode}
\usepackage{hyperref}
\usepackage{subcaption}
\usepackage{booktabs,tabularx}
\usepackage{enumitem}
\newcommand{\minisection}[1]{\noindent{\textbf{#1}}}

\newcommand{\revise}[1]{\textcolor{black}{#1}}
\pgfplotsset{compat=1.5}

\makeatletter
\newcommand{\printfnsymbol}[1]{%
  \textsuperscript{\@fnsymbol{#1}}%
}
\makeatother

\ifCLASSOPTIONcompsoc
  \usepackage[nocompress]{cite}
\else
  \usepackage{cite}
\fi

\ifCLASSINFOpdf
\else
\fi

\begin{document}
%
\title{Real-Time Scene Text Detection\\ with Differentiable Binarization\\ and Adaptive Scale Fusion}
%
%
%
%

\author{
Minghui Liao,
Zhisheng Zou,
Zhaoyi Wan,
Cong Yao,
Xiang Bai
\IEEEcompsocitemizethanks{\IEEEcompsocthanksitem 
This work was done when M. Liao and Z. Zou were with the School of Electronic Information and Communications, Huazhong University of Science and Technology, Wuhan, 430074, China. 
\IEEEcompsocthanksitem M. Liao is with Huawei Cloud, Shenzhen, 518129, China.
\IEEEcompsocthanksitem Z. Zou is with WeRide, Wuhan, 430070, China.
\IEEEcompsocthanksitem Z. Wan is with University of Rochester, Rochester, 14627, US.
\IEEEcompsocthanksitem C. Yao is with Alibaba DAMO Academy, Beijing, 100102, China.
\IEEEcompsocthanksitem X. Bai is with the School of Artificial Intelligence and Automation, Huazhong University of Science and Technology, Wuhan, 430074, China.

}
\thanks{Corresponding author: Xiang Bai.}
}

\IEEEtitleabstractindextext{%

\begin{abstract}
Recently, segmentation-based scene text detection methods have drawn extensive attention in the scene text detection field, because of their superiority in detecting the text instances of arbitrary shapes and extreme aspect ratios, profiting from the pixel-level descriptions. However, the vast majority of the existing segmentation-based approaches are limited to their complex post-processing algorithms and the scale robustness of their segmentation models, where the post-processing algorithms are not only isolated to the model optimization but also time-consuming and the scale robustness is usually strengthened by fusing multi-scale feature maps directly.
In this paper, we propose a Differentiable Binarization (DB) module that integrates the binarization process, one of the most important steps in the post-processing procedure, into a segmentation network. Optimized along with the proposed DB module, the segmentation network can produce more accurate results, which enhances the accuracy of text detection with a simple pipeline. Furthermore, an efficient Adaptive Scale Fusion (ASF) module is proposed to improve the scale robustness by fusing features of different scales adaptively. By incorporating the proposed DB and ASF with the segmentation network, our proposed scene text detector consistently achieves state-of-the-art results, in terms of both detection accuracy and speed, on five standard benchmarks.
\end{abstract}

}

\maketitle

\IEEEdisplaynontitleabstractindextext

%
\IEEEpeerreviewmaketitle

\IEEEraisesectionheading{\section{Introduction}\label{sec:introduction}}

\IEEEPARstart{R}{eading} text in scene images~\cite{long2020scene} is of great importance in both academia and industry due to its abundant real-world applications, including office automation, visual search, geo-location, and blind auxiliary. 
Scene text detection, which aims to localize the text instances in the images, is an essential component in scene text reading. Although huge progress has been achieved in recent years, scene text detection is still challenging due to the diverse scales, irregular shapes, and extreme aspect ratios of the text instances. 

As a mainstream of scene text detection, segmentation-based scene text detectors usually have advantages in handling text instances of irregular shapes and extreme aspect ratios due to their pixel-level representation and local prediction.
However, most of them rely on complex post-processing algorithms to group the pixels into text regions, resulting in a considerable time cost in the inference period. For instance, PSENet~\cite{wang2019shape} applied a progressive scale expansion algorithm to integrate multi-scale results and Tian \textit{et al.}~\cite{tian2019learning} adopted pixel embedding to group the pixels by calculating the feature distances among pixels. Besides, they mostly boosted the scale robustness of the segmentation network by applying a feature-pyramid~\cite{fpn} or U-Net~\cite{ronneberger2015u} structure to fuse the feature maps of different scales, which did not explicitly fuse the multi-scale features adaptively for the text instances of different scales.

\begin{figure}[!t]
\centering
\includegraphics[width=0.98\linewidth]{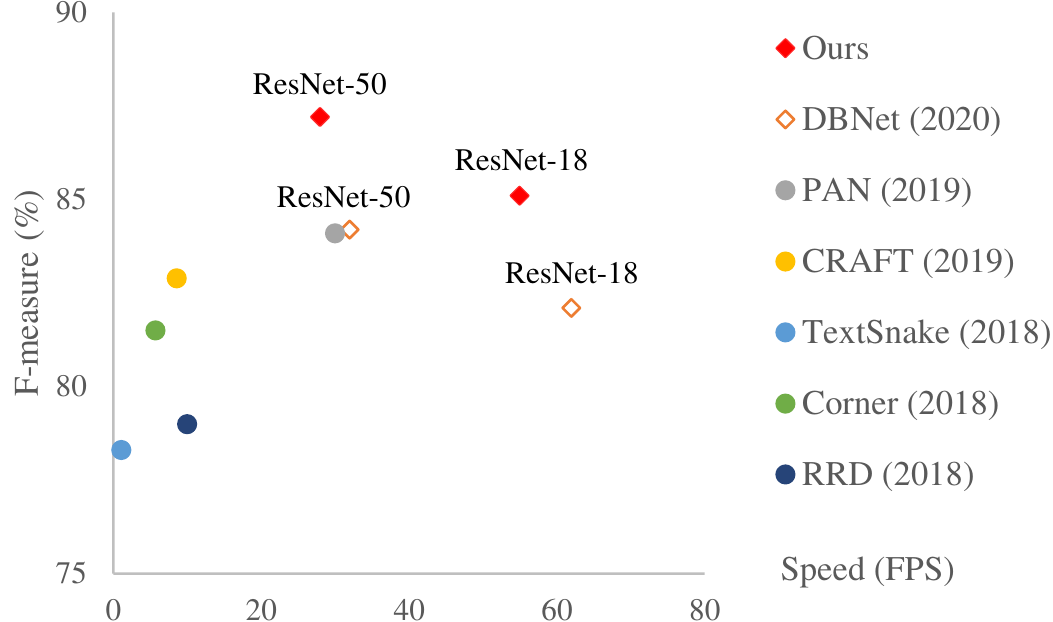}
\caption{The comparisons of several recent scene text detection methods on the MSRA-TD500 dataset~\cite{MSRA}, in terms of both accuracy (F-measure) and speed. Our method achieves the ideal tradeoff between effectiveness and efficiency.}
\label{fig:performance_speed}
\end{figure}

\begin{figure}[htbp]
\centering
\includegraphics[width=0.98\linewidth]{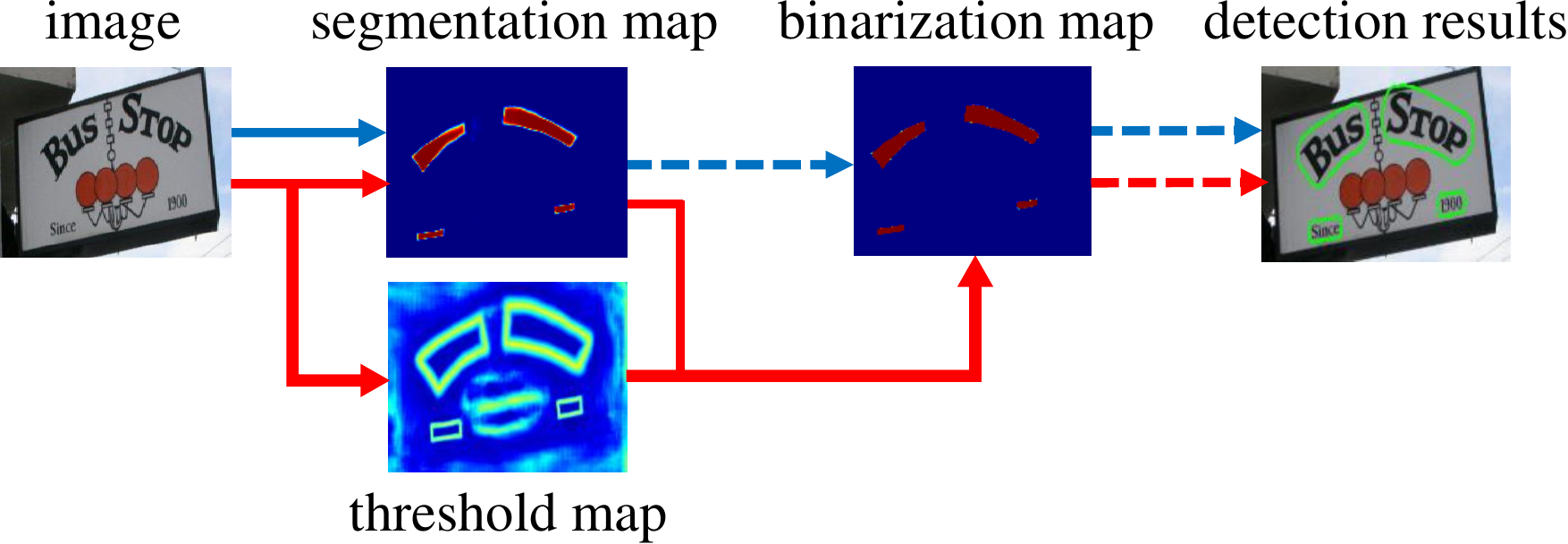}
\caption{A traditional pipeline (blue flow) and our pipeline (red flow). Dashed arrows are the inference only operators; solid arrows indicate differentiable operators in both training and inference.}
\label{fig:post-processing}
\end{figure}

A basic post-processing pipeline is described in Fig.~\ref{fig:post-processing} (following the blue arrows): First, the probability map produced from the segmentation network is converted to a binary image by applying a step function with a constant threshold; Then, some heuristic techniques like pixel clustering are used to group pixels into text regions. The above-mentioned two processes are standalone, without participating in the training process of the segmentation network, which may cause low detection accuracy.
For more effective post-processing procedures while keeping the efficiency, we propose to insert the binarization operation into the segmentation network for joint optimization (following the red arrows
in Fig.~\ref{fig:post-processing}).
First, a threshold map is predicted adaptively, where the thresholds can be diverse in different regions. This design is inspired by the observation that the boundary regions of the text instances may be with lower confidence scores than the central regions in the segmentation results. Then, we introduce an approximate function for binarization called Differentiable Binarization (DB), which binarizes the segmentation map using the threshold map. In this manner, the segmentation network is jointly optimized with the binarization process, leading to better detection results.

Different from the previous methods that directly fused the multi-scale feature maps to improve the scale robustness of the segmentation network, we propose an Adaptive Scale Fusion (ASF) module to adaptively fuse the multi-scale feature maps. ASF integrates a spatial attention module into a stage-wise attention module. The stage-wise attention module learns the weights of the feature maps of different scales and the spatial attention module learns the attention across the spatial dimensions, leading to scale-robust feature fusion.

This paper is an extension of its conference version DBNet~\cite{LiaoWYCB20}. It extends the conference version from two aspects. First, it proposes an ASF module to further enhance the scale robustness of the segmentation model, without obvious loss of efficiency. The improvements brought by the proposed ASF are positively related to the scale distribution of the scene text benchmarks.
Second, we give a more comprehensive theoretical analysis for the proposed DB module. 

An accurate, robust, and efficient scene text detector, named DBNet++, is created by integrating the proposed DB module and ASF module into a segmentation network. Profiting from the joint optimization with the segmentation network, DB not only improves the quality of the segmentation results but also contributes to a simpler post-processing algorithm. By applying ASF to the segmentation network, its ability to detect the text instances of diverse scales is distinctly strengthened. The prominent advantages of DBNet++ over the previous state-of-the-art methods are concluded as follows:
\begin{enumerate}[itemsep=0ex]
    \item Jointly optimized with the proposed DB module, our segmentation network can produce highly robust segmentation results, significantly improving the text detection results.
    \item As DB can be removed in the inference period without sacrificing the accuracy, there is no extra memory/time cost for inference. 
    \item The scale robustness of the segmentation model can be efficiently improved by the proposed ASF module.
    \item DBNet++ achieves consistently state-of-the-art accuracy on five scene text detection benchmarks, including horizontal, multi-oriented, and curved text.
\end{enumerate}

The rest paper is organized as follows: Sec.~\ref{sec:related-work} reviews the relevant scene text detection methods. We describe DBNet++ in Sec.~\ref{sec:methodology}. The experiments are discussed and analyzed in Sec.~\ref{sec:experiments}. The conclusions are summarized in Sec.~\ref{sec:conclusion}.


\section{Related Work}\label{sec:related-work}
\subsection{Text Detection}
The early scene text detection methods usually detected the individual characters or components and then grouped them into words. Neumann and Matas~\cite{neumann2012real} proposed to locate characters by classifying Extremal Regions (ERs). They posed the character detection problem as an efficient sequential selection from the set of Extremal Regions. Then, the detected ERs were grouped into words. Jaderberg \textit{et al.}~\cite{jaderberg2016reading} first generated word candidates with a proposal generator. Then, the word candidates were filtered by a random forest classifier. Finally, the remaining word candidates were refined with a regression network.

Recently, deep learning has dominated the scene text detection area. The deep-learning-based scene text detection methods can be roughly classified into three categories according to the granularity of the predicted target: regression-based methods, part-based methods, and segmentation-based methods.

\textit{Regression-based methods}
are a series of models which directly regress the bounding boxes of the text instances. TextBoxes~\cite{LiaoSBWL17} modified the anchors and the scale of the convolutional kernels based on SSD~\cite{liu2015ssd} for text detection. TextBoxes++~\cite{TextBoxes++} and DMPNet~\cite{deepmatch} applied quadrilaterals regression to detect multi-oriented text. SSTD~\cite{sstd} proposed an attention mechanism to roughly identifies text regions. RRD~\cite{liao2018rotation} decoupled the classification and regression by using rotation-invariant features for classification and rotation-sensitive features for regression, for better effect on multi-oriented and long text instances. EAST~\cite{east} and DeepReg~\cite{deepdirect} are anchor-free methods, which applied pixel-level regression for multi-oriented text instances.  DeRPN~\cite{xie2019derpn} proposed a dimension-decomposition region proposal network to handle the scale problem in scene text detection.
Regression-based methods usually enjoy simple post-processing algorithms (e.g. non-maximum suppression). However, most of them are limited to represent accurate bounding boxes for irregular shapes, such as curved shapes.

\textit{Part-based methods} firstly detect the small parts of the text instances and then link/combine them into word/text-line bounding boxes. SegLink~\cite{seglink} regressed the bounding boxes of the text segment and predicted their links, to deal with long text instances. SegLink++~\cite{tang2019seglink++} further proposed an instance-aware component grouping algorithm to separate the close text instances more effectively and improved the linking algorithm to fit the arbitrary-shape text instances. These methods usually are skilled at detecting long text lines. However, the linking algorithms are quite complex with hand-crafted super-parameters, which makes them hard to tune.

\textit{Segmentation-based methods} 
usually combine pixel-level prediction and post-processing algorithms to get the bounding boxes. Zhang \textit{et al.}~\cite{Zhang_2016_CVPR} detected multi-oriented text by semantic segmentation and MSER-based algorithms. Text border is used in Xue \textit{et al.}~\cite{xue2018accurate} to split the text instances.
Mask TextSpotter~\cite{lyu2018mask,liao2019mask} detected arbitrary-shape text instances in an instance segmentation manner based on Mask R-CNN~\cite{he2017mask}. PSENet~\cite{wang2019shape} proposed progressive scale expansion by segmenting the text instances with different scale kernels. Pixel embedding is proposed in Tian \textit{et al.}~\cite{tian2019learning} to cluster the pixels from the segmentation results. 
PSENet~\cite{wang2019shape} and SAE~\cite{tian2019learning} proposed new post-processing algorithms for the segmentation results, resulting in lower inference speed. Instead, our method focuses on improving the segmentation results by including the binarization process into the training period, without the loss of the inference speed.

Fast scene text detection methods
focus on both the accuracy and the inference speed. TextBoxes~\cite{LiaoSBWL17}, TextBoxes++~\cite{TextBoxes++}, SegLink~\cite{seglink}, and RRD~\cite{liao2018rotation} achieved fast text detection by following the detection architecture of SSD~\cite{liu2015ssd}. EAST~\cite{east} proposed to use an anchor-free design to achieve a good tradeoff between accuracy and speed. Most of them can not deal with text instances of irregular shapes, such as curved shapes. Compared to the previous fast scene text detectors, our method not only runs faster but also can detect text instances of arbitrary shapes.
Recently, PAN~\cite{wang2019efficient} proposed to adopt a low computational-cost segmentation head
and learnable post-processing for scene text detection, yielding an efficient and accurate arbitrary-shaped text detector. Our proposed DBNet++ performs more accurately and more efficiently owing to the simple and efficient differentiable binarization algorithm.

\subsection{Attention Mechanisms for Image Classification}
Some previous image classification methods use channel attention and spatial attention to enhance the accuracy of image classification. Wang \textit{et al.}~\cite{wang2017residual} proposed residual attention network for image classification. It adopted a mixture of channel attention and spatial attention to produce a soft mask for the features. Hu \textit{et al.}~\cite{hu2018squeeze} proposed a ``Squeeze-and-Excitation" block that recalibrates channel-wise feature responses by explicitly modeling interdependencies between channels. Woo \textit{et al.}~\cite{woo2018cbam} proposed to adopt a channel attention module and a spatial attention module to refine the features. These methods mainly refine the independent features by various types of attention. Our proposed adaptive scale fusion focuses on fusing the features of different scales.

\subsection{Multi-Scale Feature Fusion and Context Enhancement for Semantic Segmentation}
Context is critical for semantic segmentation methods. Context and scale are two highly related concepts, where the context can help perceive large-scale objects/scenes while the multi-scale fusion strategies can usually provide more context.
Thus, multi-scale feature fusion is commonly applied in semantic segmentation methods.

\minisection{Multi-Scale Feature Fusion}
FCN~\cite{long2015fully} firstly proposed the fully convolutional network to fuse multi-scale features with upsampling layers. U-net~\cite{ronneberger2015u} applied the skip connection which directly connected the low-level features and high-level features while inherited the structure from the FCN. PSPNet~\cite{zhao2017pyramid} and Deeplabv3 ~\cite{chen2017rethinking} proposed a pyramid pooling module (PPM) and an Atrous Spatial Pyramid Pooling (ASPP) for multi-scale feature fusion respectively.
RefineNet~\cite{lin2017refinenet}, Deeplabv3+~\cite{chen2018encoder}, DFN~\cite{yu2018learning}, and SPGNet~\cite{cheng2019spgnet} adopted encoder-decoder structure which fuse high-level and low-level features to get more discriminating feature. 
Compared to these multi-scale feature fusion methods, our proposed ASF learns the weights of multi-scale features along with the attention across both the scale and the spatial dimensions.

\minisection{Context Enhancement with Attention Mechanisms}
Attention mechanisms are popular in the semantic segmentation methods for enhancing the context. ANN~\cite{zhu2019asymmetric}, CCNet~\cite{Huang_2019_ICCV}, GCNet~\cite{cao2019gcnet}, DANet~\cite{fu2019dual} and ACFNet~\cite{zhang2019acfnet} all used self-attention to fuse different features for contextual information. Choi \textit{et al.}~\cite{choi2020cars} used height-driven attention to get height-dimensional information from multi-scale feature. Compared to these methods that applied attention mechanisms to the spatial dimensions to enhance the context, our proposed ASF focuses on the attentional fusion of multi-scale features.

\section{Methodology}\label{sec:methodology}
\begin{figure*}[ht]
\begin{center}
\includegraphics[width=0.98\linewidth]{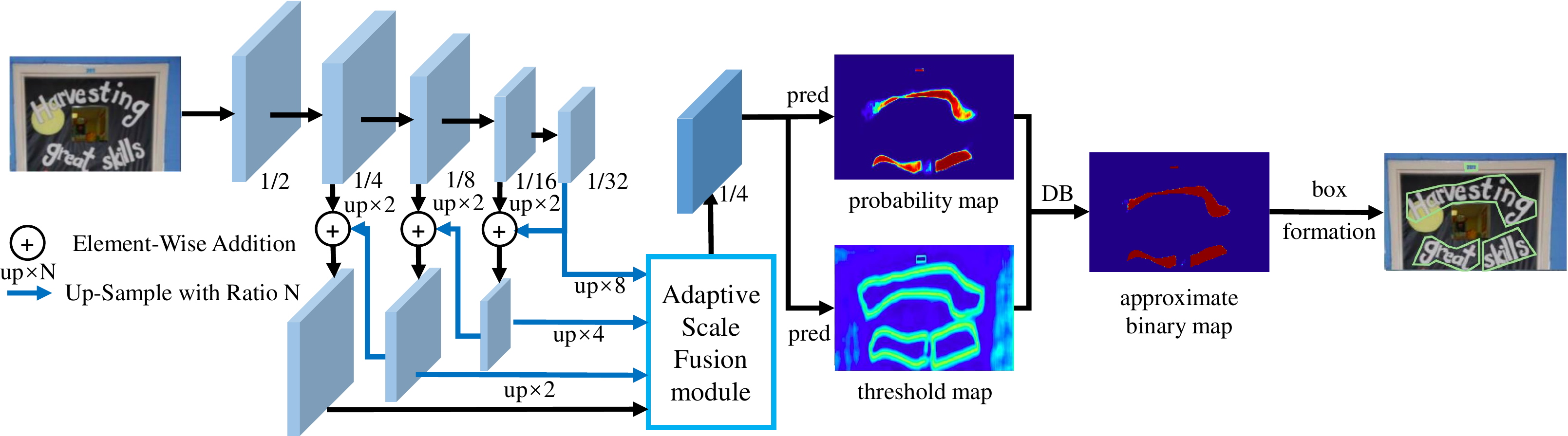}
\end{center}
\caption{Architecture of our proposed DBNet++, where the Adaptive Scale Fusion module is shown in Fig.~\ref{fig:ASF}.}
\label{fig:pipeline}
\end{figure*}

The architecture of our proposed method is shown in Fig.~\ref{fig:pipeline}. Firstly, the input image is fed into a feature-pyramid backbone. 
Secondly, the pyramid features are up-sampled to the same scale and passed to the Adaptive Scale Fusion (ASF) module to produce contextual feature $F$. Then, feature $F$ is used to predict both the probability map ($P$) and the threshold map ($T$). After that, the approximate binary map ($\hat{B}$) is calculated by $P$ and $F$. In the training period, the supervision is applied on the probability map, the threshold map, and the approximate binary map, where the probability map and the approximate binary map share the same supervision. In the inference period, the bounding boxes can be obtained easily from the approximate binary map or the probability map by a box formation process.

\begin{figure*}[ht]
\begin{center}
\includegraphics[width=0.9\linewidth]{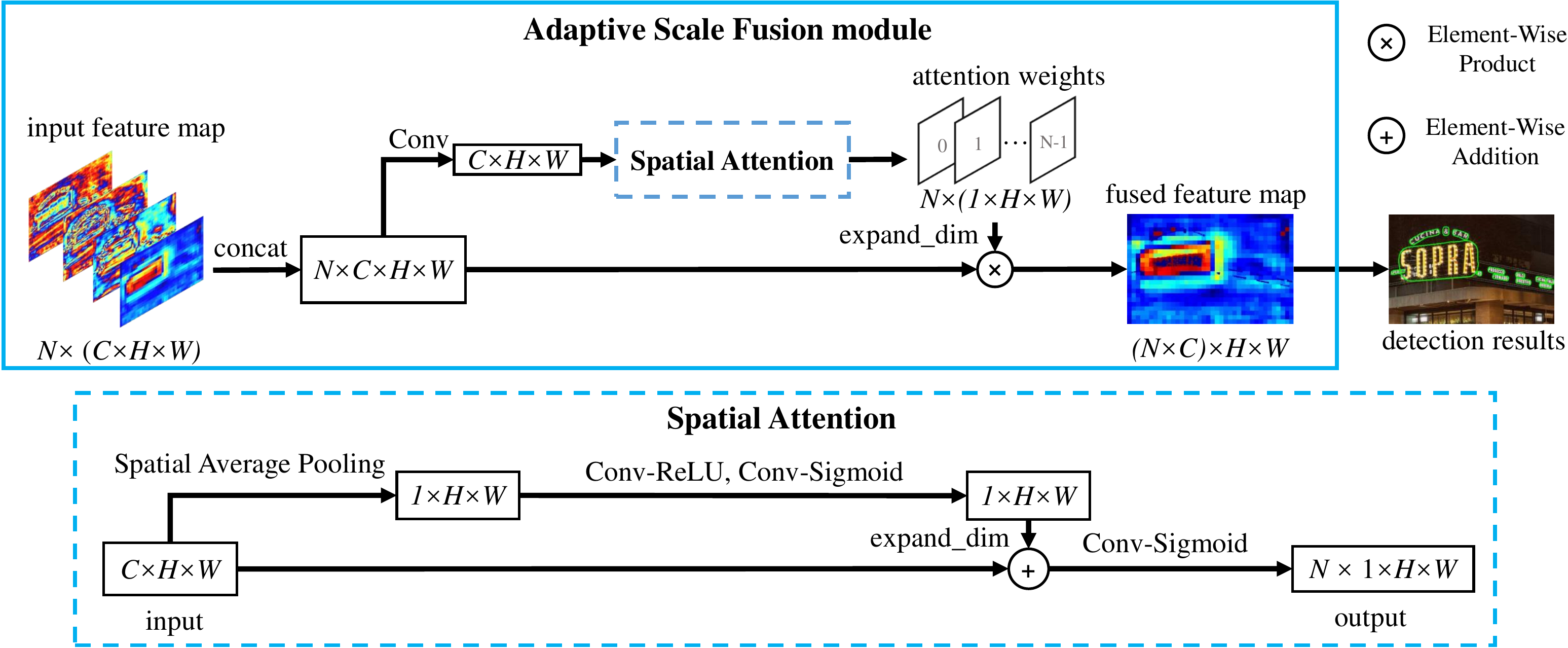}
\end{center}
\caption{Illustration of the Adaptive Scale Fusion module.}
\label{fig:ASF}
\end{figure*}

\subsection{Adaptive Scale Fusion}
The features of different scales are with different perceptions and receptive fields, thus they focus on describing the text instances of different scales. For example, the shallow or large-size features can perceive details of the small text instances but can not capture a global view of the large text instances while the deep or small-size features are opposite. To take full advantage of the features of different scales, feature-pyramid~\cite{fpn} or U-Net~\cite{ronneberger2015u} structures are commonly adopted in semantic segmentation methods. Different from most of the semantic segmentation methods that fuse the features of different scales by simply cascading or summing up, our proposed Adaptive Scale Fusion is designed to dynamically fuse the features of different scales.

As shown in Fig.~\ref{fig:ASF}, the features of different scales are scaled into the same resolution before being fed into the ASF module. Assuming that the input feature maps consist of $N$ feature maps $X \in \mathcal{R}^{N \times C \times H \times W}=\{X_{i}\}^{N-1}_{i=0}$, where N is set to 4.
Firstly, we concatenate the scaled input features $X$ and then a $3\times3$ convolutional layer is followed to obtain an intermediate feature $S \in \mathcal{R}^{C \times H \times W}$. Secondly, the attention weights $A \in \mathcal{R}^{N \times H \times W}$ can be calculated by applying a spatial attention module to the feature $S$. Thirdly, the attention weights $A$ can be split into $N$ parts along the channel dimension and weighted multiply with corresponding scaled feature to get the fused feature $F \in \mathcal{R}^{N \times C \times H \times W}$. In this way, the scale attention is defined as:
\begin{equation}
    \begin{aligned}
    S &= Conv(concat([X_{0}, X_{1}, ... , X_{N-1}])) \\
    A &= Spatial\_Attention(S)\\
    F &= concat([E_{0}X_{0}, E_{1}X_{1}, ... , E_{N-1}X_{N-1}]) \\
    \end{aligned}
    \label{eq:scale_attention}
\end{equation}
where $concat$ indicates the concatenation operator; $Conv$ represents the $3\times3$ convolutional operator; $Spatial\_Attention$ indicates a spatial attention module, which is illustrated in Fig.~\ref{fig:ASF}.
The spatial attention mechanism in the ASF makes the attention weights more flexible across the spatial dimension.

\subsection{Binarization}


\begin{figure*}[htbp]
\centering
\captionsetup[subfigure]{justification=centering}
\begin{subfigure}[b]{0.32\textwidth}
         \centering
         \includegraphics[width=\textwidth]{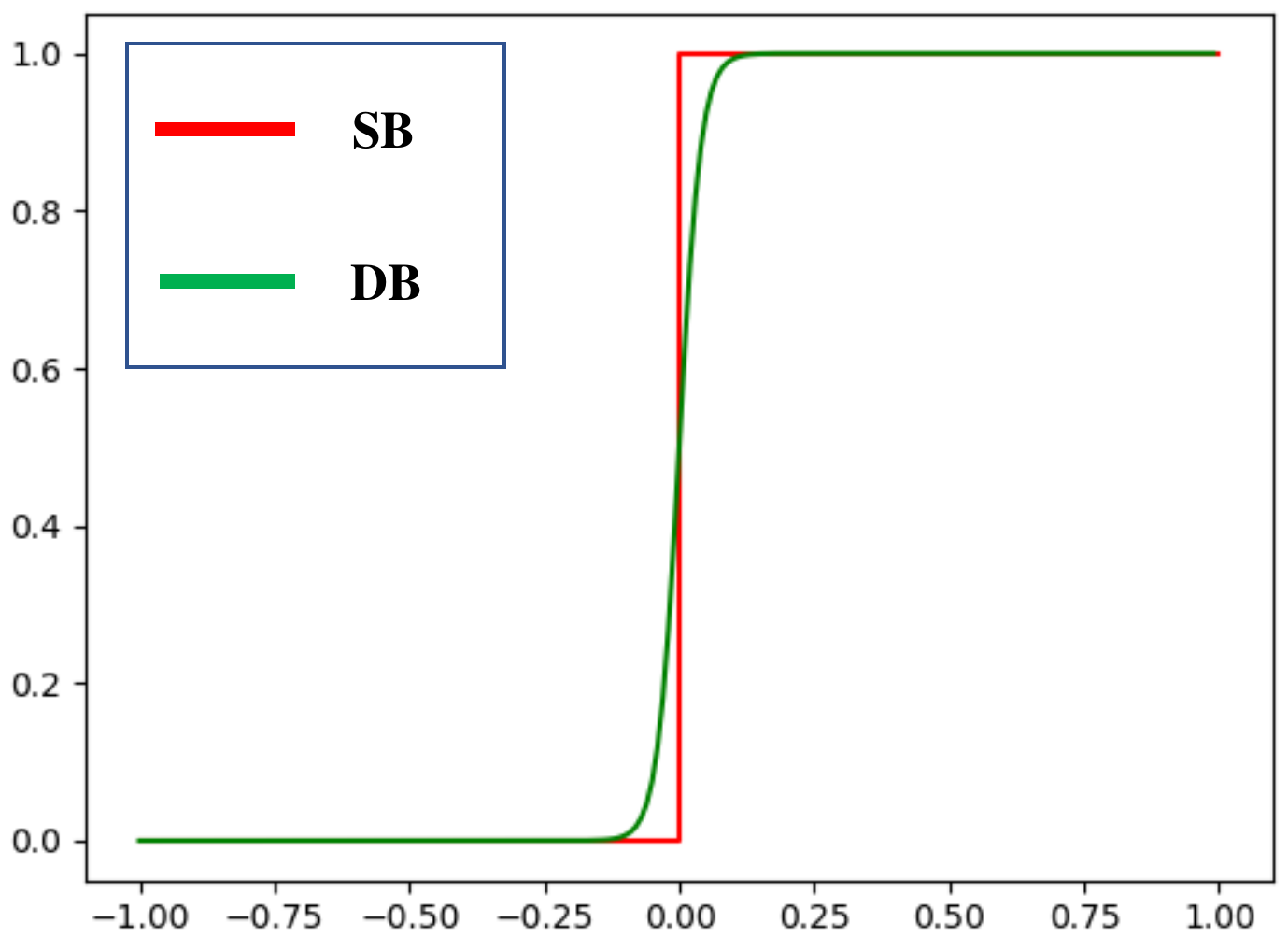}
         \caption{SB and DB}
         \label{fig:step_function}
\end{subfigure}
\begin{subfigure}[b]{0.32\textwidth}
         \centering
         \includegraphics[width=\textwidth]{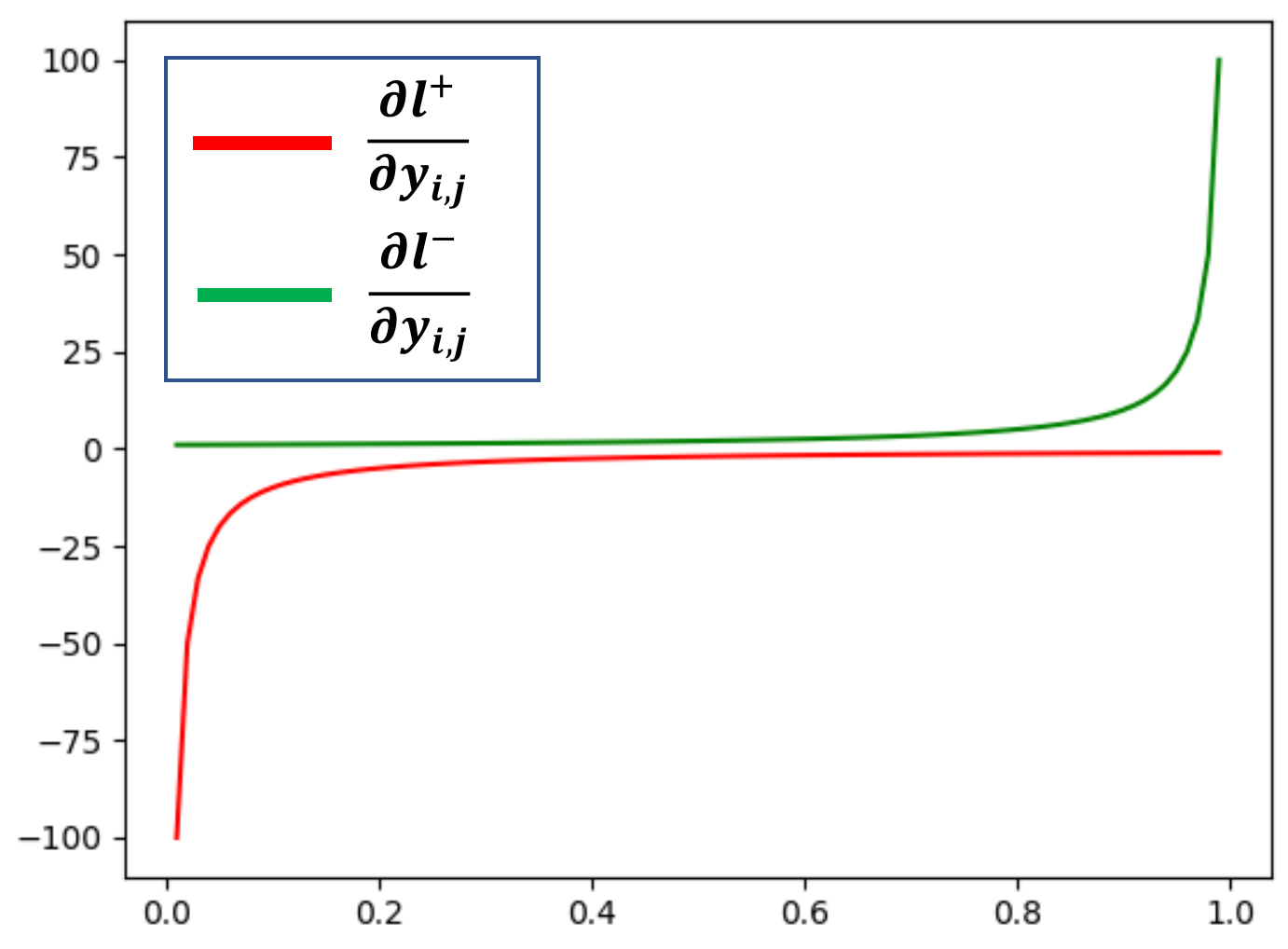}
         \caption{The derivative of losses in Eq.~\ref{Eq:derivatives_baseline}}
         \label{fig:function1}
\end{subfigure}
\begin{subfigure}[b]{0.32\textwidth}
         \centering
         \includegraphics[width=\textwidth]{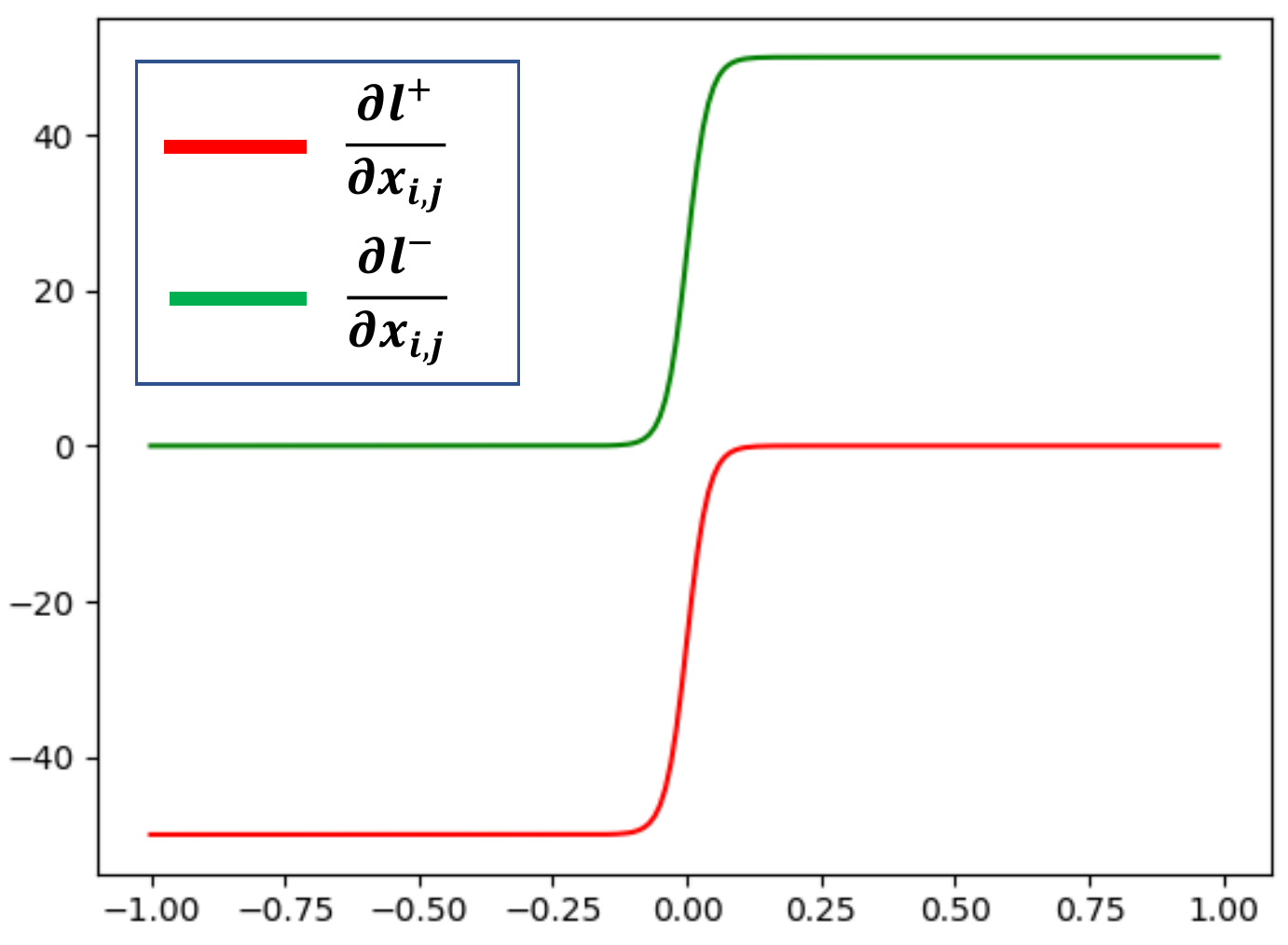}
         \caption{The derivative of losses in Eq.~\ref{Eq:derivatives_db}}
         \label{fig:function2}
\end{subfigure}
\caption{Numerical comparisons of different functions and derivatives.}
\label{fig:functions}
\end{figure*}

\noindent\textbf{Standard Binarization}
Given a probability map $P \in R^{H \times W}$ produced by a segmentation network, where $H$ and $W$ indicate the height and width of the map, it is essential to convert it to a binary map $B \in R^{H \times W}$, where pixels with value $1$ are considered as valid text areas. Usually, this binarization process can be described as follows:
\begin{equation}
    B_{i, j}=
    \begin{cases}
        1& \text{if } P_{i, j} \geq t, \\
        0&  \text{otherwise.}
    \end{cases}
    \label{eq:binarization}
\end{equation}
where $t$ is the predefined threshold and $(i, j)$ indicates the coordinate point in the map.

\noindent\textbf{Differentiable Binarization}
The standard binarization described in Eq.~\ref{eq:binarization} is not differentiable. Thus, it can not be optimized along with the segmentation network in the training period. To solve this problem, we propose to perform binarization with an approximate step function:
\begin{equation}
    \hat{B}_{i, j}=\frac{1}{1+e^{-k (P_{i, j} - T_{i, j})}}
    \label{eq:db}
\end{equation}
where $\hat{B}$ is the approximate binary map; $T$ is the adaptive threshold map learned from the network; $k$ indicates the amplifying factor. $k$ is set to $50$ empirically. 
This approximate binarization function behaves similar to the standard binarization function (see Fig~\ref{fig:step_function}) but is differentiable thus can be optimized along with the segmentation network in the training period.
The differentiable binarization with adaptive thresholds can not only help differentiate text regions from the background, but also separate text instances which are closely jointed. 
Some examples are illustrated in Fig.\ref{fig:visu}.

\subsection{\revise{Analysis of Differentiable Binarization}}
The reasons that DB improves the performance can be explained by the gradients in the backpropagation. 
Let’s take the binary cross-entropy loss as an example. The binary cross-entropy loss can be expressed as:
\begin{equation}
L_{bce} = -\frac{1}{N}\sum_{i=1}^{H}\sum_{j=1}^{W}\hat{y_{i,j}}log(y_{i,j})+(1-\hat{y_{i,j}})log(1-y_{i,j})
\end{equation}
where $y_{i,j} \in [0,1]$ and $\hat{y_{i,j}} \in \{0,1\}$ indicate the output value with logits and the target value. 
Thus, in the segmentation task, the loss $l^{+}$ for positive labels and $l^{-}$ for negative labels are:
\begin{equation}
    \begin{split}
    l^{+} &= - \log (y_{i,j}) \\
    l^{-} &= - \log (1 - y_{i,j})
    \end{split}
\end{equation}

\noindent\textbf{\revise{Without Considering Activation Function}} The differential of the segmentation loss can be calculated with the chain rule:
\begin{equation}
    \begin{split}
        \frac {\partial l^{+}}{\partial y_{i,j}} &= \frac{-1}{y_{i,j}}\\
        \frac {\partial l^{-}}{\partial y_{i,j}} &= \frac{1}{1-y_{i,j}}
    \end{split}
    \label{Eq:derivatives_baseline}
\end{equation}
Let $x_{i,j} = P_{i, j} - T_{i, j}$. The DB function can be expressed as $f(x) = \frac {1} {1 + e^{-kx_{i,j}}}$.
Similarly, the loss $l_b^{+}$ for positive labels and $l_b^{-}$ for negative labels are:
\begin{equation}
    \begin{split}
    l_b^{+} &= - \log \frac {1}{1+e^{-kx_{i,j}}} \\
    l_b^{-} &= - \log ( 1 - \frac {1}{1+e^{-kx_{i,j}}} )
    \end{split}
\end{equation}
The differential of the losses with the DB function are as follows:
\begin{equation}
    \begin{split}
        \frac {\partial l_b^{+}}{\partial x_{i,j}} &= \frac{-ke^{-kx_{i,j}}}{1+e^{-kx_{i,j}}}\\
        \frac {\partial l_b^{-}}{\partial x_{i,j}} &= \frac{k}{1+e^{-kx_{i,j}}}
    \end{split}
    \label{Eq:derivatives_db}
\end{equation}
The numerical comparison of the derivatives of the losses in Eq.~\ref{Eq:derivatives_baseline} and Eq.~\ref{Eq:derivatives_db} are also shown in Fig.~\ref{fig:function1} and Fig.~\ref{fig:function2} respectively, from which we can perceive:

(1) The magnitude of the differential around the boundary value. For the standard binary cross-entropy loss with logits (top), the magnitudes of $\frac {\partial l^{+}}{\partial y_{i,j}}$ and $\frac {\partial l^{-}}{\partial y_{i,j}}$ are quite small around the boundary value (0.5) between positive value ($>0.5$) and negative value ($<0.5$). As a result, the backpropagation or the feedback may not be significant when a predicted value is ambiguous, such as 0.4 or 0.6; For the binary cross-entropy with differentiable binarization (bottom), the magnitudes of $\frac {\partial l_b^{+}}{\partial x_{i,j}}$ and $\frac {\partial l_b^{-}}{\partial x_{i,j}}$ are large around the boundary value (0) between positive value ($>0$) and negative value ($<0$), where is augmented by the amplifying factor $k$. Thus, the proposed differentiable binarization helps to produce more distinctive predictions around the boundary value.

(2) The least upper bound and the greatest lower bound. For the standard binary cross-entropy loss with logits (top), there is no greatest lower bound for $\frac {\partial l^{+}}{\partial y_{i,j}}$ and no least upper bound for $\frac {\partial l^{-}}{\partial y_{i,j}}$; For the binary cross-entropy with differentiable binarization (bottom), the greatest lower bound of $\frac {\partial l_b^{+}}{\partial x_{i,j}}$ and the least upper bound of $\frac {\partial l_b^{-}}{\partial x_{i,j}}$ is determined by the amplifying factor $k$. Thus, the proposed differentiable binarization tends to work more stable for some extremely small or large values.


\begin{figure}[ht]
\begin{center}
\includegraphics[width=0.9\linewidth]{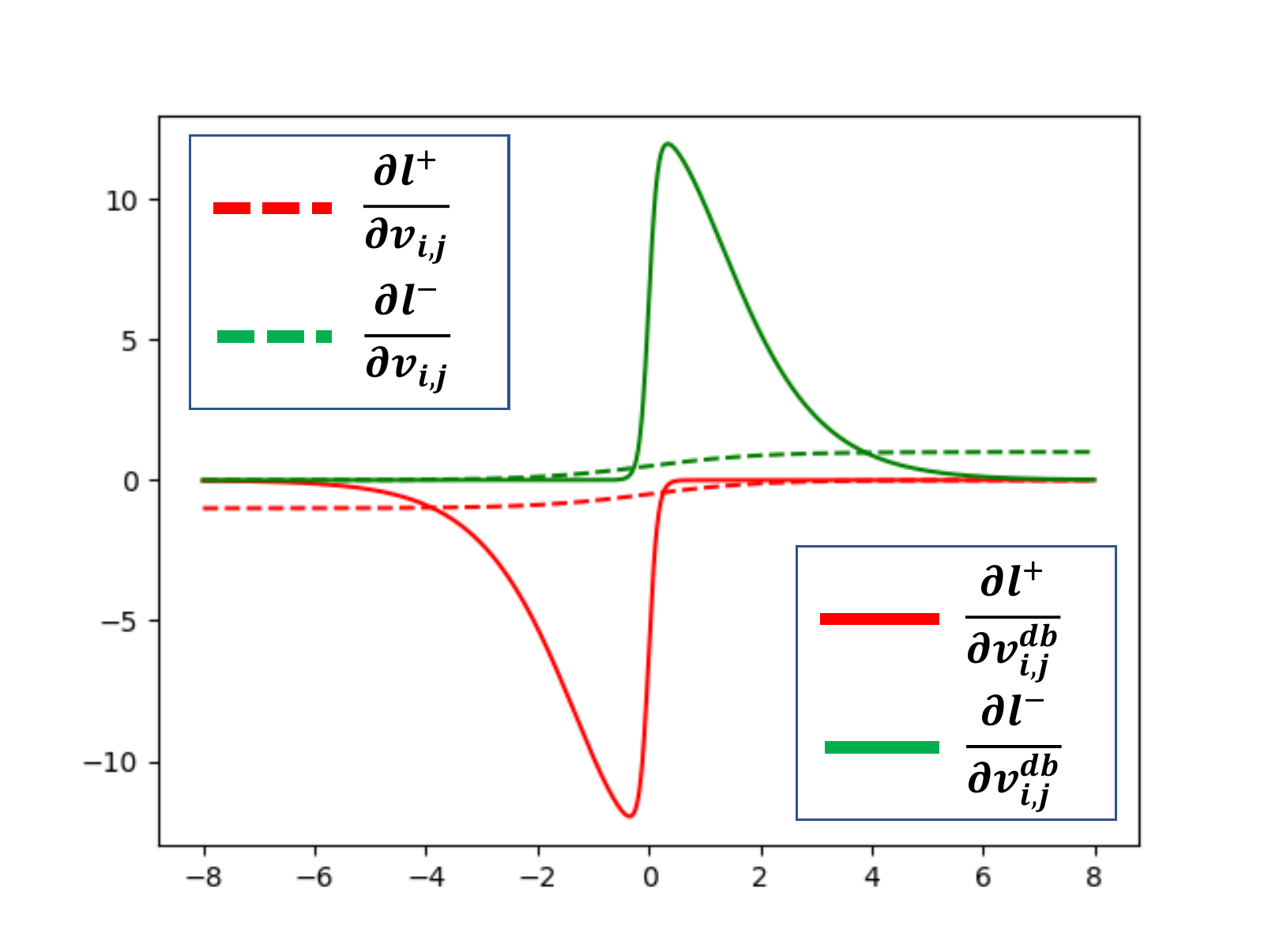}
\end{center}
\caption{\revise{The derivative of losses in Eq.~\ref{Eq:derivatives_baseline_sigmoid} and Eq.~\ref{Eq:derivatives_db_sigmoid}.}}
\label{fig:function_with_sigmoid}
\end{figure}

\noindent\textbf{\revise{Considering Activation Function}}
\revise{In practice, the activation function (Sigmoid) is applied in both formulations, which could bound the derivatives and alleviate the problem of ``The least upper bound and the greatest lower bound". Considering the Sigmoid function as follows:
\begin{equation}
    \begin{split}
    y_{i,j} &= \frac{1}{1+e^{-v_{i,j}}} \\
    x_{i,j} &= \frac{1}{1+e^{-v^{db}_{i,j}}} - T_{i,j},
    \end{split}
    \label{Eq:with_sigmoid}
\end{equation}
where $-v_{i,j}$ and $v^{db}_{i,j}$ indicate the output values before the Sigmoid activations; $T_{i,j}$ indicates the adaptive threshold. In this way, the differential of the losses can be updated by chain rule as follows:
\begin{equation}
    \begin{split}
        \frac {\partial l^{+}}{\partial v_{i,j}} &= \frac{-e^{-v_{i,j}}}{1+e^{-v_{i,j}}}\\
        \frac {\partial l^{-}}{\partial v_{i,j}} &= \frac{1}{1+e^{-v_{i,j}}}
    \end{split}
    \label{Eq:derivatives_baseline_sigmoid}
\end{equation}
\begin{equation}
    \begin{split}
        \frac {\partial l_b^{+}}{\partial v^{db}_{i,j}} &= \frac{-ke^{-k(\frac{1}{1+e^{-v^{db}_{i,j}}}-T_{i,j})-v^{db}_{i,j}}}{(1+e^{-kv^{db}_{i,j}})^2(1+e^{-k(\frac{1}{1+e^{-v^{db}_{i,j}}}-T_{i,j})})}\\
        \frac {\partial l_b^{-}}{\partial v^{db}_{i,j}} &= \frac{k}{1+e^{-kv^{db}_{i,j}}}
    \end{split}
    \label{Eq:derivatives_db_sigmoid}
\end{equation}
Without loss of generality, Eq.~\ref{Eq:derivatives_baseline_sigmoid} and Eq.~\ref{Eq:derivatives_db_sigmoid} can be visualized as Fig.~\ref{fig:function_with_sigmoid}, by setting $k=50$ and $T_{i,j}=0.5$. 
}

\revise{
We can perceive from Fig.~\ref{fig:function_with_sigmoid} that the differentiable binarization enlarges the feedback of backpropagation when the wrongly predicted values are near the boundary value. Thus, the proposed differentiable binarization makes the model focus on optimizing the prediction of ambiguous regions. Besides, the Sigmoid function alleviates the problem of ``The least upper bound and the greatest lower bound" and DB further decreases the penalty for extremely small/large values.
}

\subsection{Adaptive Threshold}
\begin{figure}[htbp]
\captionsetup[subfigure]{justification=centering}
    \begin{subfigure}{0.453\linewidth}
    \centering
        \includegraphics[width=\linewidth]{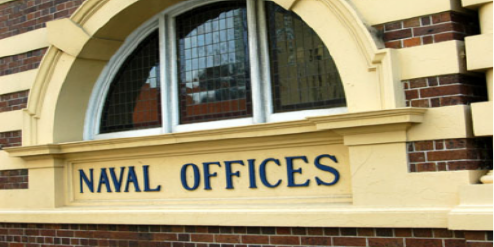}
        \caption{Input image}
    \end{subfigure} %
    \qquad
    \begin{subfigure}{0.453\linewidth}
    \centering
        \includegraphics[width=\linewidth]{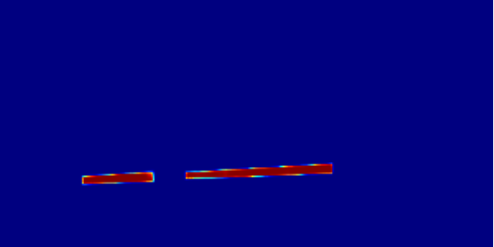}
        \caption{Probability map}
   \end{subfigure}

    \begin{subfigure}{0.453\linewidth}
    \centering
        \includegraphics[width=\linewidth]{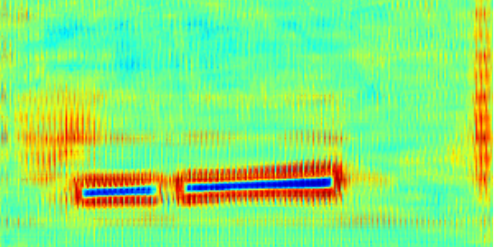}
        \caption{Threshold map without supervision}
    \end{subfigure} %
    \qquad
    \begin{subfigure}{0.453\linewidth}
    \centering
        \includegraphics[width=\linewidth]{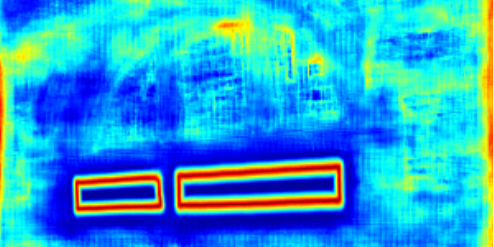}
        \caption{Threshold map with supervision}
   \end{subfigure}
\caption{The threshold map with/without supervision. 
}
\label{fig:supervision}
\end{figure}

The threshold map in Fig.~\ref{fig:performance_speed} is similar to the text border map in \cite{xue2018accurate} from appearance. However, the motivation and usage of the threshold map are different from the text border map. The threshold map with/without supervision is visualized in Fig.~\ref{fig:supervision}. 
The threshold map would highlight the text border region even without supervision for the threshold map. This indicates that the border-like threshold map is beneficial to the final results. Thus, we apply border-like supervision on the threshold map for better guidance. An ablation study about the supervision of the adaptive threshold is discussed in the Experiments section.
For the usage, the text border map in \cite{xue2018accurate} is used to split the text instances while our threshold map is served as thresholds for the binarization. 

\subsection{Deformable Convolution}
Deformable convolution~\cite{dai2017deformable,zhu2019deformable} can provide a flexible receptive field for the model, which is especially beneficial to the text instances of extreme aspect ratios. 
Following~\cite{zhu2019deformable}, modulated deformable convolutions are applied in all the $3 \times 3$ convolutional layers in stages conv3, conv4, and conv5 in the ResNet-18 or ResNet-50 backbone~\cite{he2016deep}.

\subsection{Label Generation}
\begin{figure}[htbp]
\centering
\includegraphics[width=0.95\linewidth]{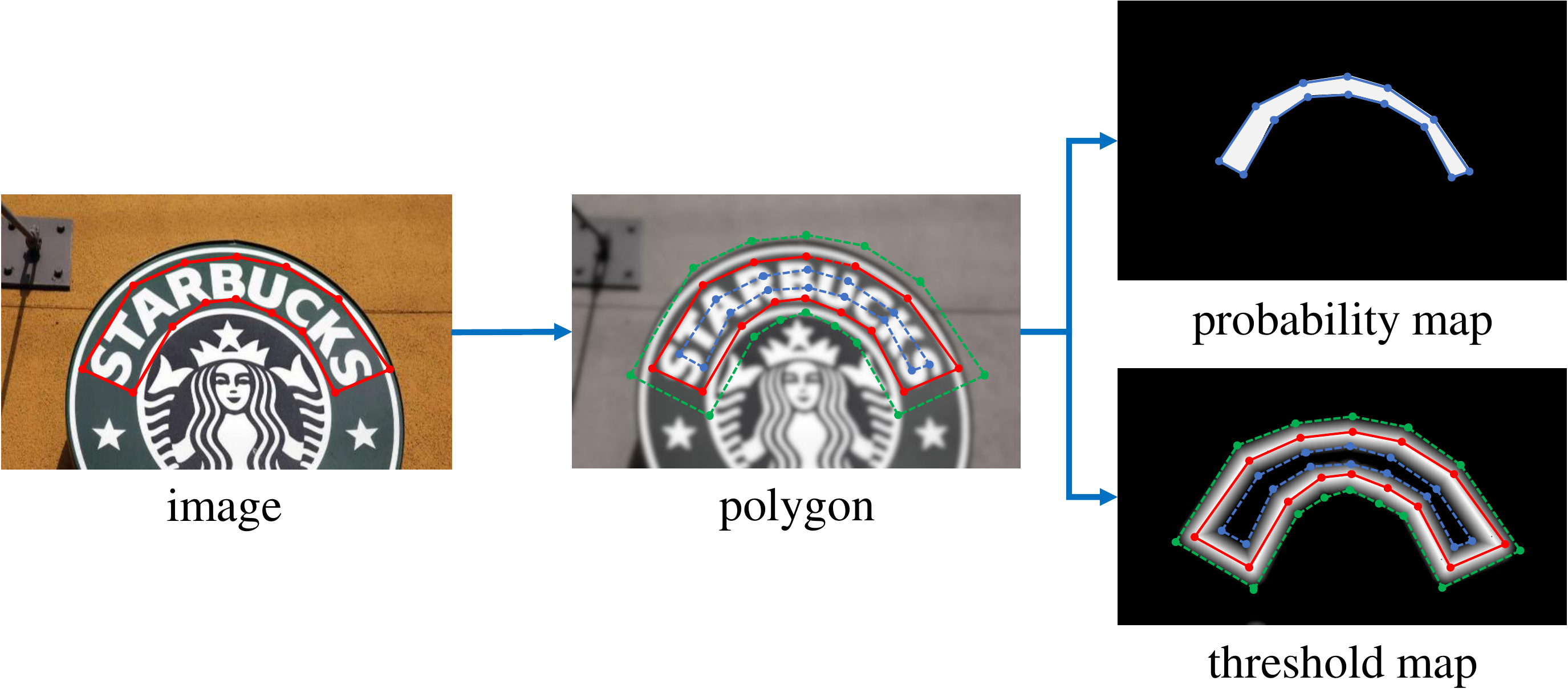}
\caption{Label generation. The annotation of text polygon is visualized in red lines. The shrunk and dilated polygon are displayed in blue and green lines, respectively.}
\label{fig:label}
\end{figure}

The label generation for the probability map is inspired by PSENet~\cite{wang2019shape}. 
Given a text image, each polygon of its text regions is described by a set of segments:
\begin{equation}
    G = \{S_{k}\}^{n}_{k=1}
\end{equation}
$n$ is the number of vertexes, which may be different in different datasets, e.g, 4 for the ICDAR 2015 dataset~\cite{icdar15} and 16 for the CTW1500 dataset~\cite{ctw1500}. Then the positive area is generated by shrinking the polygon $G$ to $G_{s}$ using the Vatti clipping algorithm~\cite{vati}. The offset $D$ of shrinking is computed from the perimeter $L$ and area $A$ of the original polygon: 
\begin{equation}
    D = \frac{A(1-r^{2})}{L}
\end{equation}
where $r$ is the shrink ratio, set to $0.4$ empirically.

With a similar procedure, we can generate labels for the threshold map. Firstly the text polygon $G$ is dilated with the same offset $D$ to $G_{d}$. We consider the gap between $G_{s}$ and $G_{d}$ as the border of the text regions, where the label of the threshold map can be generated by computing the distance to the closest segment in $G$.

\subsection{Optimization}
The loss function $L$ can be expressed as a weighted sum of the loss for the probability map $L_{s}$, the loss for the binary map $L_{b}$, and the loss for the threshold map $L_t$:
\begin{equation}
    L = L_{s} + \alpha \times L_{b} + \beta \times L_{t}
\end{equation}
According to the numeric values of the losses, $\alpha$ and $\beta$ are set to $1.0$ and $10$ respectively.

We apply a binary cross-entropy (BCE) loss for both $L_s$ and $L_b$. To overcome the unbalance of the number of positives and negatives, hard negative mining is used in the BCE loss by sampling the hard negatives.
\begin{equation}
    L_s = L_b = \sum_{i \in S_{l}}{y_i}\log{x_i} + (1-y_i)\log{(1-x_i)}
\end{equation}
$S_{l}$ is the sampled set where the ratio of positives and negatives is $1:3$. 
\revise{It consists of all the positives and the top-$k$ negatives (sorted by the values of the predicting probability), where $k$ is 3 times the number of positives.}

$L_{t}$ is computed as the sum of $L1$ distances between the prediction and label inside the dilated text polygon $G_d$:
\begin{equation}
    L_t = \sum_{i \in R_d}{|y^{*}_i - x^{*}_i|}
\end{equation}
where $R_d$ is a set of indexes of the pixels inside the dilated polygon $G_d$; $y^{*}$ is the label for the threshold map.

\begin{figure*}[ht]
\centering
\includegraphics[width=0.98\linewidth]{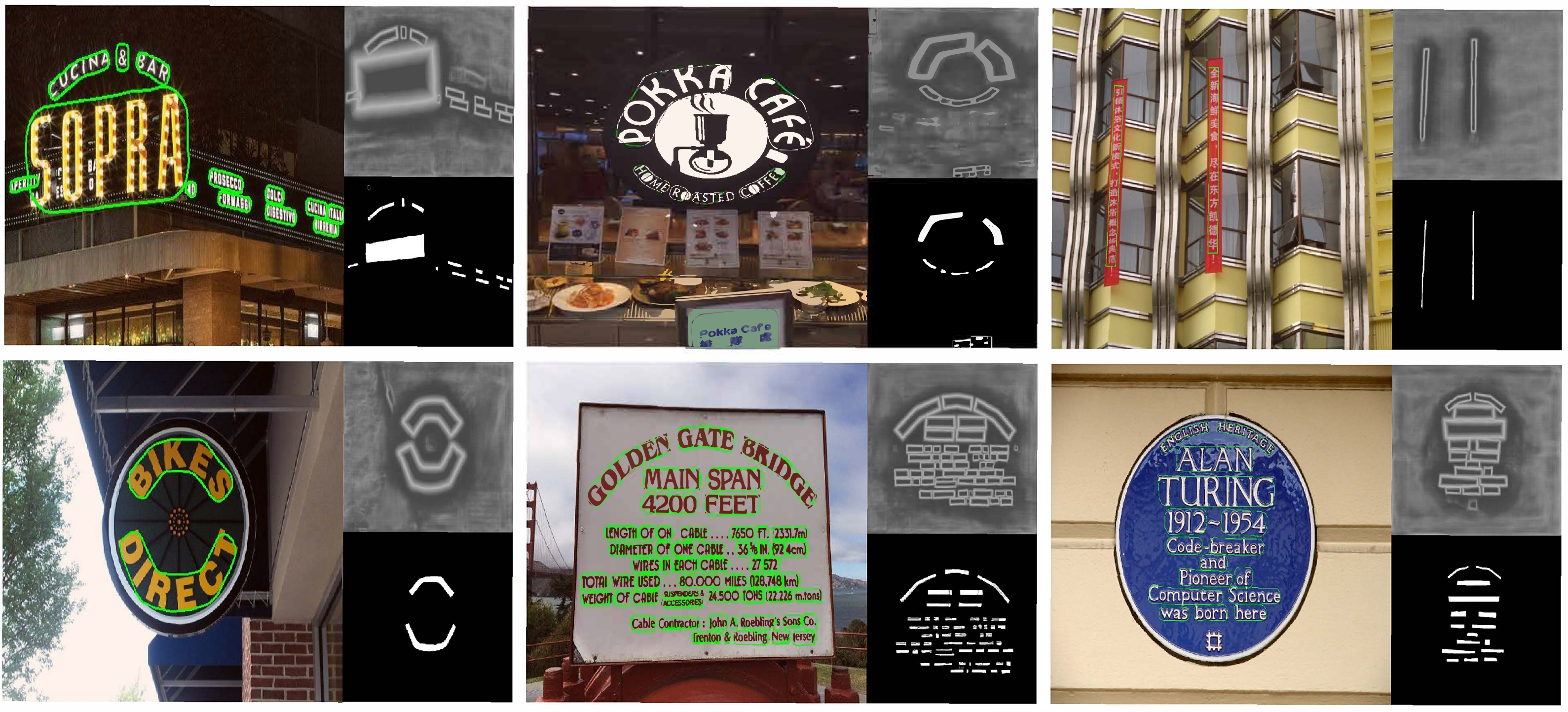}
\caption{Some visualization results on text instances of various shapes, including curved text, multi-oriented text, vertical text, and long text lines. For each unit, the top right is the threshold map; the bottom right is the probability map.}
\label{fig:visu}
\end{figure*}

In the inference period, we can either use the probability map or the approximate binary map to generate text bounding boxes, which produces almost the same results. For better efficiency, we use the probability map so that the threshold branch can be removed. The box formation process consists of three steps: (1) the probability map/the approximate binary map is firstly binarized with a constant threshold (0.2), to get the binary map; (2)the connected regions (shrunk text regions) are obtained from the binary map; (3) the shrunk regions are dilated with an offset $D'$ the Vatti clipping algorithm\cite{vati}. $D'$ is calculated as 
\begin{equation}
    D' = \frac{A' \times r'}{L'}
\end{equation}
where $A'$ is the area of the shrunk polygon; $L'$ is the perimeter of the shrunk polygon; $r'$ is set to $1.5$ empirically.

\section{Experiments}\label{sec:experiments}

\subsection{Datasets}
The scene text datasets used in the experiments are described as follows.

\textit{SynthText}~\cite{SynthText} is a synthetic dataset which consists of $800k$ images. These images are synthesized from $8k$ background images. This dataset is only used to pre-train our model.

\textit{MLT-2017 dataset}~\footnote{https://rrc.cvc.uab.es/?ch=8} is a multi-language dataset. It includes $9$ languages representing 6 different scripts. There are $7,200$ training images, $1,800$ validation images, and $9,000$ testing images in this dataset. We use both the training set and the validation set in the finetune period.

\textit{MLT-2019 dataset}~\footnote{https://rrc.cvc.uab.es/?ch=15} is a multi-language dataset. It is an extension of MLT-2017. It includes $10$ languages representing $7$ different scripts. The languages include Chinese, Japanese, Korean, English, French, Arabic, Italian, German, Bangla, and Hindi (Devanagari). There are $10,000$ training images, $2,000$ validation images, and $10,000$ testing images in this dataset. We use the training set in the finetune period. 

\textit{ICDAR 2015 dataset}~\cite{icdar15} consists of $1,000$ training images and $500$ testing images, which are captured by Google glasses with a resolution of $720 \times 1280$. The text instances are labeled at the word level. 

\textit{MSRA-TD500 dataset}~\cite{MSRA} is a multi-language dataset that includes English and Chinese. There are $300$ training images and $200$ testing images. The text instances are labeled at the text-line level. Following the previous methods~\cite{east,lyu2018multi,long2018textsnake}, we include extra $400$ training images from HUST-TR400~\cite{yao2014unified}.

\textit{CTW1500 dataset}~\cite{ctw1500} mainly focuses on curved text. It consists of $1,000$ training images and $500$ testing images. The text instances are annotated at the text-line level.

\textit{Total-Text dataset}~\cite{totaltext} includes the text of various shapes, including horizontal, multi-oriented, and curved. They are $1,255$ training images and $300$ testing images. The text instances are labeled at the word level.

\begin{table*}[ht]
\setlength{\tabcolsep}{13.0pt}
\centering
\caption{Detection results with different settings of deformable convolution, differentiable binarization and adaptive scale fusion module. ``DConv" indicates deformable convolution. ``ASF" indicates adaptive scale fusion module. ``P'', ``R'', and ``F'' indicate precision, recall, and f-measure respectively.}
\begin{tabularx}{1.0\linewidth}{lc*{10}c}
\toprule
\multirow{2}{*}{Backbone} & \multirow{2}{*}{DConv} & \multirow{2}{*}{DB} &  \multirow{2}{*}{ASF}& \multicolumn{4}{c}{MSRA-TD500} & \multicolumn{4}{c}{CTW1500} \\ \cline{5-12} 
                          &                                         &                             &                 & P      & R      & F      & FPS & P      & R      & F      & FPS \\ 
\midrule   
ResNet-18                 & $\times$                                      & $\times$                                           & $\times$                                           & 85.5   & 70.8   & 77.4   & \textbf{66} & 76.3 & 72.8 & 74.5 & \textbf{59} \\ 
ResNet-18                 & \checkmark                                     & $\times$                                           & $\times$                                           & 86.8   & 72.3   & 78.9   & 62  & 80.9 & 75.4 & 78.1 & 55 \\ 
ResNet-18                 & $\times$       & \checkmark                &  $\times$            &  87.3   & 75.8   & 81.1   & 66  & 82.4 & 76.6 & 79.4 & 59 \\ 
ResNet-18                 & $\times$       & $\times$                  &  \checkmark          &  84.9  & 78.5    & 81.6   & 53  & 83.5 & 75.9 & 79.5 & 45 \\ 
ResNet-18                 & \checkmark     & \checkmark                &  $\times$            & \textbf{90.4}   & 76.3   & 82.8   & 62 & 84.8 & 77.5 & 81.0 & 55  \\
ResNet-18                 & \checkmark     &  $\times$                 & \checkmark           & 87.1   & 79.9   & 83.3   & 55  & 86.4 & 80.8 & 83.5 & 40 \\
ResNet-18                 & \checkmark     &  \checkmark               & \checkmark           & 87.9   & \textbf{82.5}   & \textbf{85.1}  & 55  & \textbf{86.7} & \textbf{81.3} & \textbf{83.9} & 40 \\
\midrule  

ResNet-50                 & $\times$                                      & $\times$                                           & $\times$                                      & 84.6   & 73.5   & 78.7   & \textbf{40}  & 81.6 & 72.9 & 77.0 & \textbf{27} \\ 
ResNet-50                 & \checkmark                                     & $\times$                                           & $\times$                                      & 90.5   & 77.9   & 83.7   & 32 & 86.2 & 78.0 & 81.9 & 22 \\ 
ResNet-50                 & $\times$                                      & \checkmark                                          & $\times$                                      & 86.6   & 77.7   & 81.9   & 40 & 84.3 & 79.1 & 81.6 & 27 \\ 
ResNet-50                 & $\times$       & $\times$                  &  \checkmark          &  84.5  & 83.2    & 83.8   & 32  & 83.3 & 79.1 & 81.2 & 24 \\ 

ResNet-50                 & \checkmark                                     & \checkmark                                          & $\times$                                      & 91.5   & 79.2   & 84.9   & 32 & 86.9 & 80.2 & 83.4 & 22 \\ 

ResNet-50                 &  \checkmark         & $\times$                 & \checkmark         & 90.7   & \textbf{83.5}  & 86.9   & 29 & 89.2 & 81.4 & 85.1 & 21   \\
ResNet-50                 & \checkmark          & \checkmark               & \checkmark         & \textbf{91.5}   & 83.3  & \textbf{87.2}   & 29 & \textbf{87.9} & \textbf{82.8} & \textbf{85.3} & 21 \\
\bottomrule
\end{tabularx}
\label{tab:ablation}
\end{table*}

\begin{table*}[ht]
\setlength{\tabcolsep}{13.0pt}
\centering
\caption{Detection results with different settings of ASF. ``Spatial'' means spatial attention in the adaptive scale fusion module; ``Scale'' means adaptive scale fusion.}
\begin{tabularx}{1.0\linewidth}{lc*{9}c}
\toprule
%
\multirow{2}{*}{Base Method} & \multirow{2}{*}{Scale} & \multirow{2}{*}{Spatial} & \multicolumn{4}{c}{MSRA-TD500} & \multicolumn{4}{c}{CTW1500} \\ \cline{4-11} 
                          &                                         &                                              & P      & R      & F      & FPS & P      & R      & F      & FPS \\ 
\midrule   
DBNet (ResNet-50)                 & $\times$                                      & $\times$                                           & 91.5  &  79.2   &  84.9  & 32  & 86.9 &  80.2 &  83.4  & 22 \\  
DBNet (ResNet-50)                & \checkmark                                     & $\times$                                           & 92.2   & 81.8 & 86.7  & 30 & 85.4 & 83.2 & 84.3 & 21 \\ 
DBNet (ResNet-50)               & \checkmark                                     & \checkmark                                          & 91.5   & 83.3   & 87.2   & 29 & 87.9 & 82.8 & 85.3 & 21 \\
\bottomrule
\end{tabularx}
\label{tab:ablation_asf}
\end{table*}

\subsection{Implementation Details}
For all the models, we first pre-train them with the SynthText dataset for $100k$ iterations. Then, we finetune the models on the corresponding real-world datasets for $1200$ epochs. The training batch size is set to 16. We follow a “poly” learning rate
policy where the learning rate at the current iteration equals the initial learning rate
multiplying $(1 - \frac{iter}{max\_iter})^{power}$, where the initial learning rate is set to 0.007 and $power$ is $0.9$. We use a weight decay of 0.0001 and a momentum
of 0.9. The $max\_iter$ means the maximum number of iterations, which depends on the maximum epochs.

The data augmentation for the training data includes: (1) Random rotation with an angle range of $(-10^{\circ}, 10^{\circ})$; (2) Random cropping; (3) Random Flipping. All the processed images are re-sized to $640 \times 640$ for better training efficiency.

In the inference period, we keep the aspect ratio of the test images and re-size the input images by setting a suitable height for each dataset. The inference speed is tested with a batch size of $1$, with a single GTX 1080Ti GPU. The inference time cost consists of the model forward time cost and the post-processing time cost. The post-processing time cost is about $30\%$ of the inference time.

\subsection{Ablation Study}
We conduct an ablation study on the MSRA-TD500 dataset and the CTW1500 dataset to show the effectiveness of the modules including differentiable binarization, deformable convolution, and adaptive scale fusion. The detailed experimental results are shown in Tab.~\ref{tab:ablation}, Tab.~\ref{tab:ablation_asf},  and Tab.~\ref{tab:ablation_thresh}.

\minisection{Differentiable Binarization}
In Tab.~\ref{tab:ablation}, we can see that our proposed DB improves the performance significantly for both ResNet-18 and ResNet-50 on the two datasets.
For the ResNet-18 backbone, DB achieves $3.7\%$ and $4.9\%$ performance gain in terms of F-measure on the MSRA-TD500 dataset and the CTW1500 dataset. For the ResNet-50 backbone, DB brings $3.2\%$ (on the MSRA-TD500 dataset) and $4.6\%$ (on the CTW1500 dataset) improvements. Moreover, since DB can be removed in the inference period, the speed is the same as the one without DB. 

\minisection{Deformable Convolution}
As shown in Tab.~\ref{tab:ablation}, the deformable convolution can also brings $1.5\%-5.0\%$ performance gain since it provides a flexible receptive field for the backbone, with small extra time costs. For the MSRA-TD500 dataset, the deformable convolution increase the F-measure by $1.5\%$ (with ResNet-18) and $5.0\%$ (with ResNet-50). For the CTW1500 dataset, $3.6\%$ (with ResNet-18) and $4.9\%$ (with ResNet-50) improvements are achieved by the deformable convolution.

\begin{table}[!ht]
\setlength{\tabcolsep}{10.0pt}
\centering
\caption{Effect of supervising the threshold map on the MLT-2017 dataset. ``Thr-Sup'' denotes applying supervision on the threshold map.}
\begin{tabularx}{1.0\linewidth}{lc*{5}c}
\toprule

Backbone & Thr-Sup & P      & R      & F      & FPS  \\               
\midrule                     
ResNet-18                                     & $\times$                                          & 81.3   & 63.1   & 71.0   & 41 \\ 

ResNet-18                                     & \checkmark                                          & \textbf{81.9}   & \textbf{63.8}   & \textbf{71.7}   & 41 \\ \midrule  
ResNet-50                                                    & $\times$                                           & 81.5   & 64.6   & 72.1   & 19   \\ 
ResNet-50                                      & \checkmark                                          & \textbf{83.1}   & \textbf{67.9}   & \textbf{74.7}   & 19 \\ \bottomrule
\end{tabularx}
\label{tab:ablation_thresh}
\end{table}

\minisection{Backbone}
The proposed detector with the ResNet-50 backbone achieves better performance than the ResNet-18 but runs slower. Specifically, The best ResNet-50 model outperforms the best ResNet-18 model by $2.1\%$ (on the MSRA-TD500 dataset) and $2.4\%$ (on the CTW1500 dataset), with approximate double time cost.

\minisection{Supervision of Threshold Map}
Although the threshold maps with/without supervision are similar in appearance, the supervision can bring performance gain. As shown in Tab.~\ref{tab:ablation_thresh}, the supervision improves $0.7\%$ (ResNet-18) and $2.6\%$ (ResNet-50) on the MLT-2017 dataset.

\begin{table*}[ht]
\setlength{\tabcolsep}{15.0pt}
\centering
\caption{Comparisons with multi-scale feature fusion and context enhancement modules in semantic segmentation methods. ``PPM": Pyramid
Pooling Module; ``CCA": Criss-Cross Attention.}
\begin{tabularx}{1.0\linewidth}{lc*{9}c}
\toprule
%
\multirow{2}{*}{Base Method} & \multirow{2}{*}{Module} & \multicolumn{4}{c}{MSRA-TD500} & \multicolumn{4}{c}{CTW1500} \\ \cline{3-10} 
                    &                & P      & R      & F      & FPS & P  & R    & F    & FPS \\ 
\midrule   
DBNet (ResNet-50) & PPM~\cite{zhao2017pyramid}     & 91.2   & 79.7   & 85.1   & 23 & 87.0 & 79.9 & 83.3 & 16 \\ 
DBNet (ResNet-50) & CCA~\cite{Huang_2019_ICCV}   & 92.9   & 80.9   & 86.5   & 22  & 88.4 & 81.5 & 84.8 & 15 \\  
DBNet (ResNet-50) & ASF(ours)                           & 91.5   & 83.3   & \textbf{87.2}  & \textbf{29}  & 87.9 & 82.8 & \textbf{85.3} & \textbf{21} \\
\bottomrule
\end{tabularx}
\label{tab:compare_with_multi_scale}
\end{table*}

\begin{figure*}[htbp]
\centering
\includegraphics[width=0.95\linewidth]{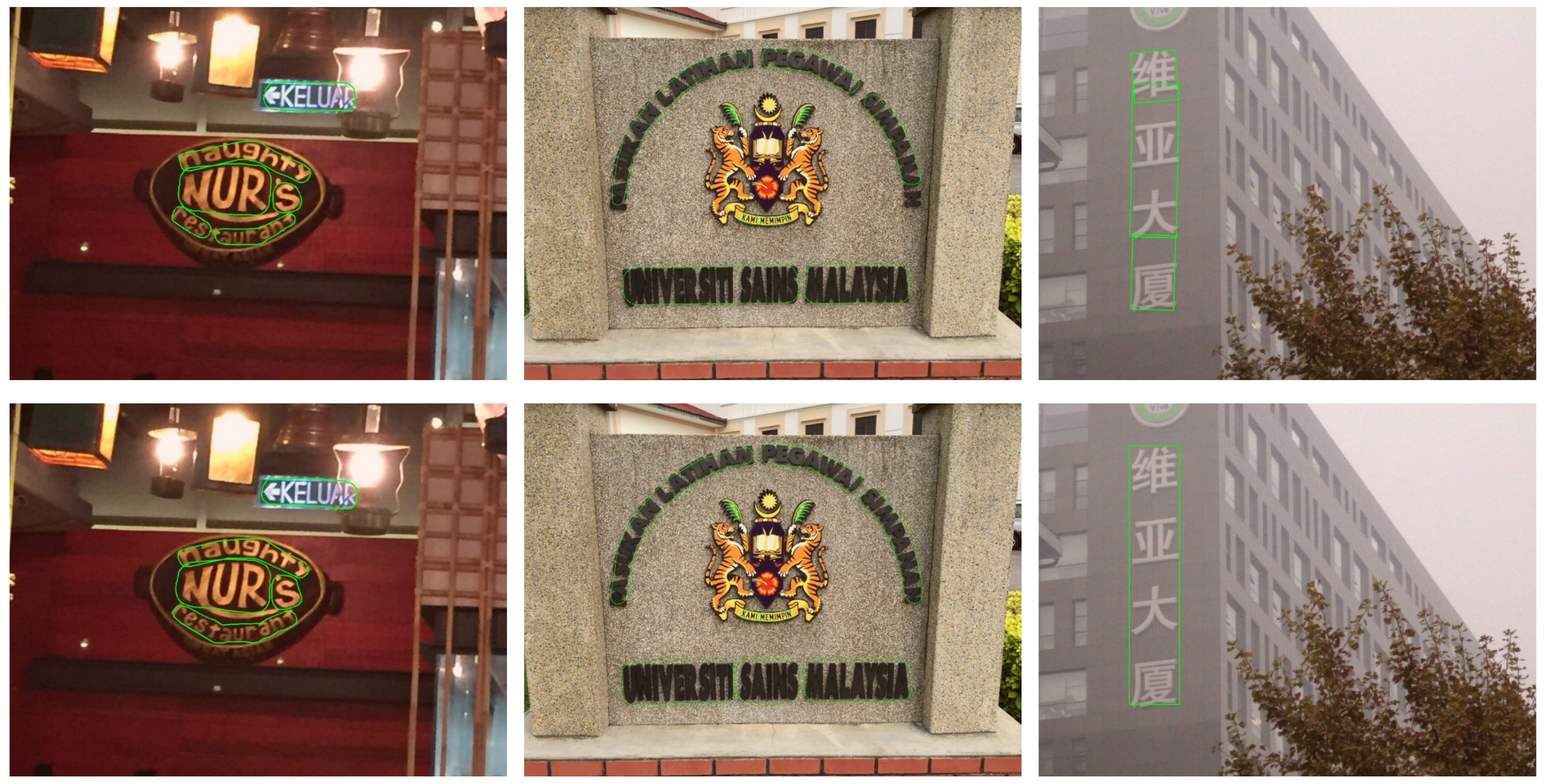}
\caption{Some visualization results of DBNet and DBNet++ on text instances of various shapes, including curved text, vertical text and multi-oriented text. For each unit, the top is the result of DBNet; the bottom is the result of DBNet++. \revise{More results are shown in the appendix.}}
\label{fig:vis_compare}
\end{figure*}

\minisection{Adaptive Scale Fusion}
As shown in Tab.~\ref{tab:ablation}, the adaptive scale fusion module improves the F-measure by $2.3\%$ and $1.9\%$ on the MSRA-TD500 dataset and the CTW1500 dataset, respectively. The inference speed decreases slightly. As shown in Tab.~\ref{tab:ablation_asf}, the spatial attention in the ASF module brings $0.5\%$ and $1.0\%$ on the MSRA-TD500 dataset and the CTW1500 dataset by providing more flexible and adaptive attention weights across the spatial dimension.

\minisection{Comparisons with PPM and CCA}
We compare our proposed ASF with a multi-scale feature fusion module (Pyramid Pooling Module, PPM~\cite{zhao2017pyramid}) and a context enhancement module (Criss-Cross Attention, CCA~\cite{Huang_2019_ICCV}). We integrate them into the encoder of the DBNet for a fair comparison.
As shown in Tab.~\ref{tab:compare_with_multi_scale}, the proposed ASF outperforms the PPM and CCA in terms of both the detection accuracy and the inference speed. The experimental results demonstrate that the proposed ASF is more effective than PPM and CCA.

Our proposed ASF performs better than PPM and CCA on the text detection task because that ASF is not designed to simply enlarge the receptive field or introduce more context for the segmentation model. It fuses the multi-scale feature maps with stage-wise and spatial-wise attention weights, which can be guided by the scales of the corresponding text regions.

\subsection{Comparisons with Previous Methods}
We compare our proposed method with previous methods on five standard benchmarks, including two benchmarks for curved text, one benchmark for multi-oriented text, and two multi-language benchmarks for long text lines. Some qualitative results are visualized in Fig.~\ref{fig:visu}.

\begin{table}[!ht]
\setlength{\tabcolsep}{10.0pt}
\centering
\caption{Detection results on the Total-Text dataset. The values in the bracket mean the height of the input images. ``*" indicates testing with multiple scales.}
\begin{tabularx}{1.0\linewidth}{lc*{3}c}
\toprule
Method           & P    & R    & F    & FPS \\ \midrule
TextSnake~\cite{long2018textsnake}      & 82.7 & 74.5 & 78.4 & -   \\ 
ATRR~\cite{atrr}            & 80.9 & 76.2 & 78.5 & -   \\ 
Mask TextSpotter~\cite{lyu2018mask} & 82.5 & 75.6 & 78.6 & -   \\ 
TextField~\cite{xu2019textfield}        & 81.2 & 79.9 & 80.6 & -   \\ 
LOMO~\cite{lomo}*        & 87.6 & 79.3 & 83.3 & -   \\ 
CRAFT~\cite{craft}           & 87.6 & 79.9 & 83.6 & -   \\ 
CSE~\cite{cse}              & 81.4 & 79.1 & 80.2 & -   \\ 
PSENet-1s~\cite{wang2019shape}        & 84.0   & 78.0   & 80.9 & 3.9 \\ 
PAN~\cite{wang2019efficient}        & \textbf{89.3}   & 81.0   & 85.0 & 39.6 \\ 
\midrule  
DBNet (ResNet-18) (800)~\cite{LiaoWYCB20}   & 88.3 & 77.9 & 82.8 & \textbf{50}  \\ 
DBNet (ResNet-50) (800)~\cite{LiaoWYCB20}  & 87.1 & 82.5 & 84.7 & 32  \\ 
\midrule
DBNet++ (ResNet-18) (800)   & 87.4 & 79.6 & 83.3 & 48  \\ 
DBNet++ (ResNet-50) (800)  & 88.9 & \textbf{83.2} & \textbf{86.0} & 28 \\ 
\bottomrule
\end{tabularx}
\label{tab:totaltext}
\end{table}

\begin{table}[!ht]
\setlength{\tabcolsep}{9.5pt}
\centering
\caption{Detection results on the CTW1500 dataset. The methods with ``*" are collected from~\cite{ctw1500}. The values in the bracket mean the height of the input images.}
\begin{tabularx}{1.0\linewidth}{lc*{3}c}
\toprule
Method        & P             & R             & F             & FPS         \\ \midrule
CTPN*          & 60.4          & 53.8          & 56.9          & 7.14        \\ 
EAST*          & 78.7          & 49.1          & 60.4          & 21.2        \\ 
SegLink*       & 42.3            & 40.0            & 40.8            & 10.7         \\ 
TextSnake~\cite{long2018textsnake}     & 67.9          & \textbf{85.3}          & 75.6          & 1.1         \\ 
TLOC~\cite{ctw1500}     & 77.4            & 69.8          & 73.4          & 13.3           \\ 
PSENet-1s~\cite{wang2019shape}    & 84.8         & 79.7          & 82.2          & 3.9         \\ 
SAE~\cite{tian2019learning}       & 82.7          & 77.8          & 80.1    & 3  \\ 
PAN~\cite{wang2019efficient}  & 86.4          & 81.2          & 83.7    & 39.8  \\ 
\midrule  
DBNet (ResNet-18) (1024)~\cite{LiaoWYCB20} & 84.8          & 77.5          & 81.0          & \textbf{55} \\ 
DBNet (ResNet-50) (1024)~\cite{LiaoWYCB20} & 86.9 & 80.2 & 83.4 & 22          \\ 
\midrule
DBNet++ (ResNet-18) (1024)   & 86.7 & 81.3 & 83.9 & 40 \\ 
DBNet++ (ResNet-18) (800)   & 84.3 & 81.0 & 82.6 & 49  \\ 
DBNet++ (ResNet-50) (1024)   & \textbf{88.5} & 82.0 & 85.1 & 21  \\ 
DBNet++ (ResNet-50) (800)  & 87.9 & 82.8 & \textbf{85.3} & 26  \\ \bottomrule
\end{tabularx}
\label{tab:CTW}
\end{table}

\minisection{Curved Text Detection}
We prove the shape robustness of our method on two curved text benchmarks (Total-Text and CTW1500). As shown in Tab.~\ref{tab:totaltext} and Tab.~\ref{tab:CTW}, our method achieves state-of-the-art performance both on accuracy and speed.
Specifically, ``DBNet++ (ResNet-50)" outperforms the previous state-of-the-art method by $1.0\%$ and $1.6\%$ on the Total-Text and the CTW1500 dataset.
``DBNet (ResNet-50)" runs faster than all previous methods and the speed can be further improved by using a ResNet-18 backbone, with a small performance drop. Compared to the recent fast text detector~\cite{wang2019efficient}, DBNet++ achieves better accuracy with a comparable inference speed.

\minisection{Multi-Oriented Text Detection}
The ICDAR 2015 dataset is a multi-oriented text dataset that contains lots of small and low-resolution text instances.
In Tab.~\ref{tab:ic15}, we can see that ``DBNet++ (ResNet-50) (1152)" and ``DBNet (ResNet-50) (1152)" achieve the state-of-the-art performance on accuracy. 
Compared to EAST~\cite{east}, ``DBNet++ (ResNet-50) (736)" outperforms it by $7.2\%$ on accuracy and runs twice faster. Compared to PAN~\cite{wang2019efficient}, ``DBNet++ (ResNet-18) (736)" performs better in terms of accuracy and inference speed. 

\begin{table}[ht]
\setlength{\tabcolsep}{9.5pt}
\centering
\caption{Detection results on the ICDAR 2015 dataset. The values in the bracket mean the height of the input images.}
\begin{tabularx}{1.0\linewidth}{lc*{3}c}
\toprule
Method        & P    & R    & F    & FPS  \\ \midrule
CTPN~\cite{eccv/TianHHH016}          & 74.2 & 51.6 & 60.9 & 7.1  \\ 
EAST~\cite{east}          & 83.6 & 73.5 & 78.2 & 13.2 \\ 
SSTD~\cite{sstd}          & 80.2 & 73.9 & 76.9 & 7.7  \\ 
WordSup~\cite{hu2017wordsup}       & 79.3 & 77   & 78.2 & -  \\ 
Lyu \textit{et al.}~\cite{lyu2018multi}    & \textbf{94.1} & 70.7 & 80.7 & 3.6  \\ 
TextBoxes++~\cite{TextBoxes++}   & 87.2 & 76.7 & 81.7 & 11.6 \\ 
RRD~\cite{liao2018rotation}           & 85.6 & 79   & 82.2 & 6.5  \\ 
MCN~\cite{mcn}           & 72   & 80   & 76   & -  \\ 
TextSnake~\cite{long2018textsnake}     & 84.9 & 80.4 & 82.6 & 1.1  \\ 
PSENet-1s~\cite{wang2019shape}    & 86.9 & 84.5 & 85.7 & 1.6  \\ 
SPCNet~\cite{spc}       & 88.7 & \textbf{85.8} & 87.2 & -  \\ 
LOMO~\cite{lomo}      & 91.3 & 83.5 & 87.2 & -  \\ 
CDAFT~\cite{craft}      & 89.8 & 84.3 & 86.9 & -  \\ 
SAE(720)~\cite{tian2019learning}  & 85.1  & 84.5  & 84.8  & 3       \\ 
SAE(990)~\cite{tian2019learning}  & 88.3   & 85.0  & 86.6  & -       \\ 
PAN~\cite{wang2019efficient}  & 84.0   & 81.9  & 82.9  & 26.1       \\ 
\midrule  
DBNet (ResNet-18) (736)~\cite{LiaoWYCB20} & 86.8 & 78.4 & 82.3 & \textbf{48}   \\ 
DBNet (ResNet-50) (1152)~\cite{LiaoWYCB20} & 91.8   & 83.2 & \textbf{87.3} & 12   \\ 
\midrule  
DBNet++ (ResNet-18) (736) & 90.1 & 77.2 & 83.1 & 44  \\ 
DBNet++ (ResNet-50) (1152) & 90.9 & 83.9 & \textbf{87.3} & 10   \\ 
\bottomrule
\end{tabularx}
\label{tab:ic15}
\end{table}

\minisection{Multi-Language Text Detection}
Our method is robust on multi-language text detection. As shown in Tab.~\ref{tab:td500} and Tab.~\ref{tab:mlt19}, ``DBNet++ (ResNet-50)" is superior to previous methods on accuracy and speed. For the accuracy, ``DBNet++ (ResNet-50)" surpasses the previous state-of-the-art method by $3.1\%$ and $3.3\%$ on the MSRA-TD500 dataset and the MLT-2019 dataset respectively. For the speed, ``DBNet++ (ResNet-18) (736)" is faster than the previous fastest method~\cite{wang2019efficient} while achieving better accuracy on the MSRA-TD500 dataset. The speed can be further accelerated to 80 FPS (``DBNet++ (ResNet-18) (512)") by decreasing the input size.

\begin{table}[ht]
\setlength{\tabcolsep}{10.0pt}
\centering
\caption{Detection results on the MSRA-TD500 dataset. The values in the bracket mean the height of the input images.}
\begin{tabularx}{1.0\linewidth}{lc*{3}c}
\toprule
Method        & P             & R             & F             & FPS         \\ \midrule
He \textit{et al.}~\cite{he2016text}          & 71            & 61            & 69            & -         \\ 
DeepReg~\cite{deepdirect}       & 77            & 70            & 74            & 1.1         \\ 
RRPN~\cite{rrpn}          & 82            & 68            & 74            & -         \\ 
RRD~\cite{liao2018rotation}           & 87            & 73            & 79            & 10          \\ 
MCN~\cite{mcn}          & 88            & 79            & 83            & -         \\ 
PixelLink~\cite{deng2018pixellink}     & 83            & 73.2          & 77.8          & 3           \\ 
Lyu \textit{et al.}~\cite{lyu2018multi}    & 87.6          & 76.2          & 81.5          & 5.7         \\ 
TextSnake~\cite{long2018textsnake}     & 83.2          & 73.9          & 78.3          & 1.1         \\ 
Xue \textit{et al.}~\cite{xue2018accurate}         & 83.0          & 77.4          & 80.1          & -       \\ 
MSR~\cite{MSR}     & 87.4          & 76.7          & 81.7          & -       \\   
CRAFT~\cite{craft}         & 88.2          & 78.2          & 82.9          & 8.6       \\ 
SAE~\cite{tian2019learning}        & 84.2          & 81.7          & 82.9          & -       \\ 
PAN~\cite{wang2019efficient}        & 84.4          & \textbf{83.8}          & 84.1          & 30.2       \\ 
\midrule  
DBNet (ResNet-18) (512)~\cite{LiaoWYCB20} & 85.7          & 73.2          & 79.0          & \textbf{82} \\ 
DBNet (ResNet-18) (736)~\cite{LiaoWYCB20} & 90.4          & 76.3          & 82.8          & 62 \\ 
DBNet (ResNet-50) (736)~\cite{LiaoWYCB20} & \textbf{91.5} & 79.2 & 84.9 & 32          \\  
\midrule  
DBNet++ (ResNet-18) (512) & 89.7   & 76.5   & 82.6  & 80 \\ 
DBNet++ (ResNet-18) (736) & 87.9   & 82.5   & 85.1  & 55 \\ 
DBNet++ (ResNet-50) (736) & \textbf{91.5}   & 83.3   & \textbf{87.2}   & 29  \\  
\bottomrule
\end{tabularx}
\label{tab:td500}
\end{table}

\begin{table}[!ht]
\setlength{\tabcolsep}{12.0pt}
\centering
\caption{Detection results on the MLT-2019 dataset. *CRAFTS used character-level annotations and integrated a recognition model.}
\begin{tabularx}{1.0\linewidth}{ll*{3}c}
\toprule
Method        & P             & R             & F             & FPS         \\ \midrule
PSENet~\cite{wang2019shape}          & 73.5            & 59.6            & 65.8            & -         \\ 
CRAFTS*~\cite{baek2020character}          & 79.5          & 59.6          & 68.1          & -        \\ 
\midrule  
DBNet (ResNet-18)~\cite{LiaoWYCB20} & 75.3          & 60.2          & 66.9          & \textbf{19} \\ 
DBNet (ResNet-50)~\cite{LiaoWYCB20} & 78.3 & 64.0 & 70.4 & 10 \\ 
\midrule  
DBNet++ (ResNet-18) & 77.5    & 61.0    & 68.2     & 18 \\ 
DBNet++ (ResNet-50) & \textbf{78.6} & \textbf{65.4}  & \textbf{71.4} & 10 \\ 
\bottomrule
\end{tabularx}
\label{tab:mlt19}
\end{table}


\begin{figure*}[htbp]
\centering
    \begin{subfigure}{0.21\linewidth}
    \centering
        \includegraphics[width=\linewidth]{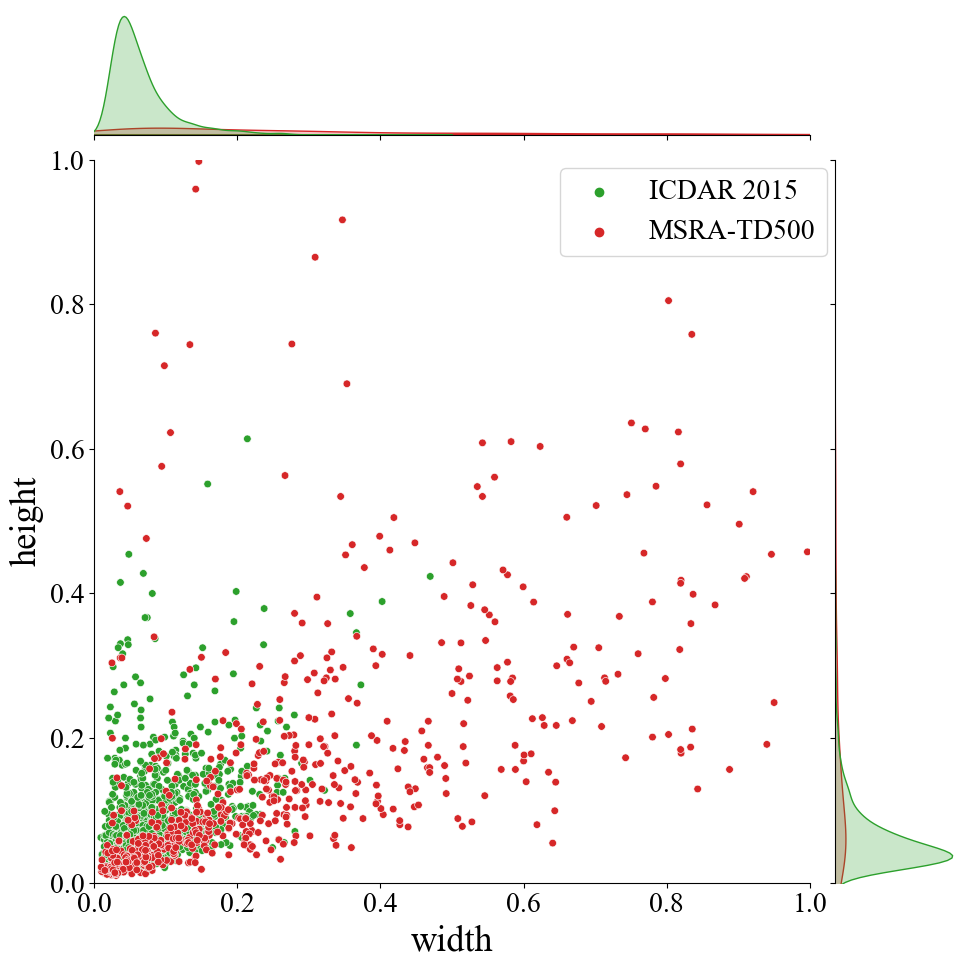}
        \caption{ICDAR 2015 dataset and MSRA-TD500 dataset}
    \end{subfigure} %
    \qquad
    \begin{subfigure}{0.21\linewidth}
    \centering
        \includegraphics[width=\linewidth]{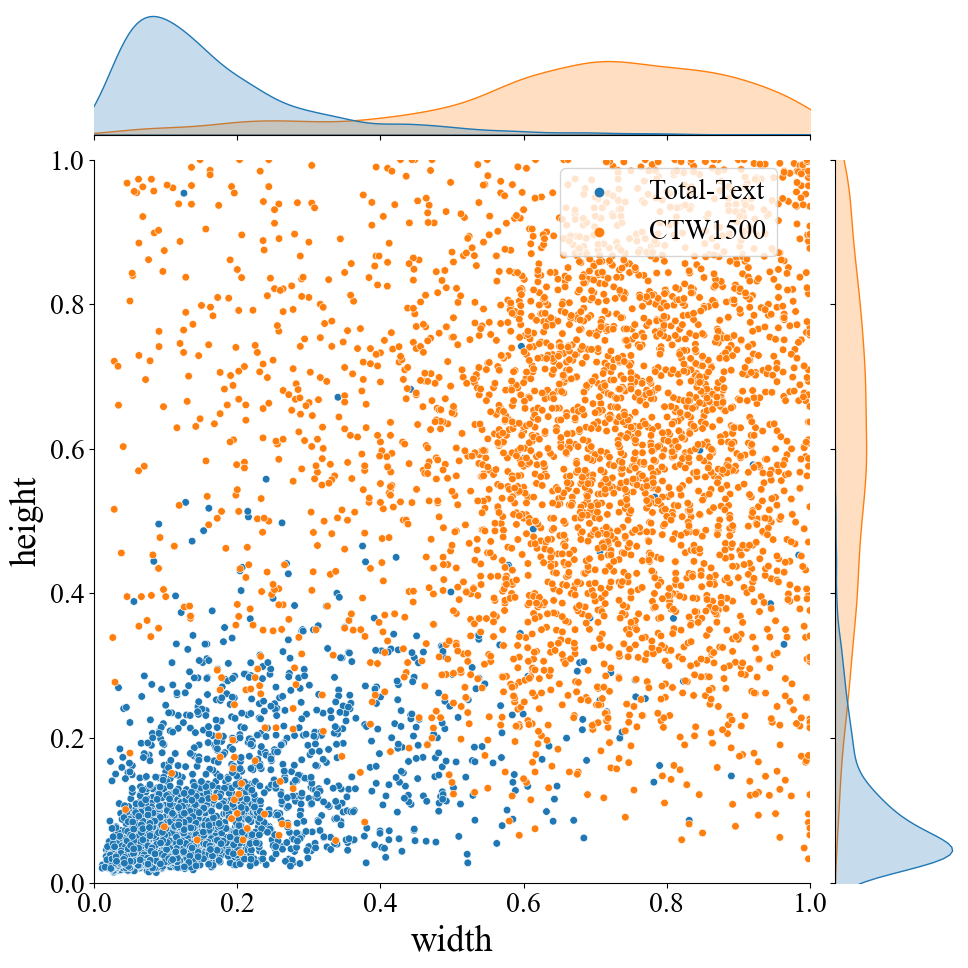}
        \caption{Total-Text dataset and CTW1500 dataset}
   \end{subfigure} %
   \qquad
    \begin{subfigure}{0.21\linewidth}
    \centering
        \includegraphics[width=\linewidth]{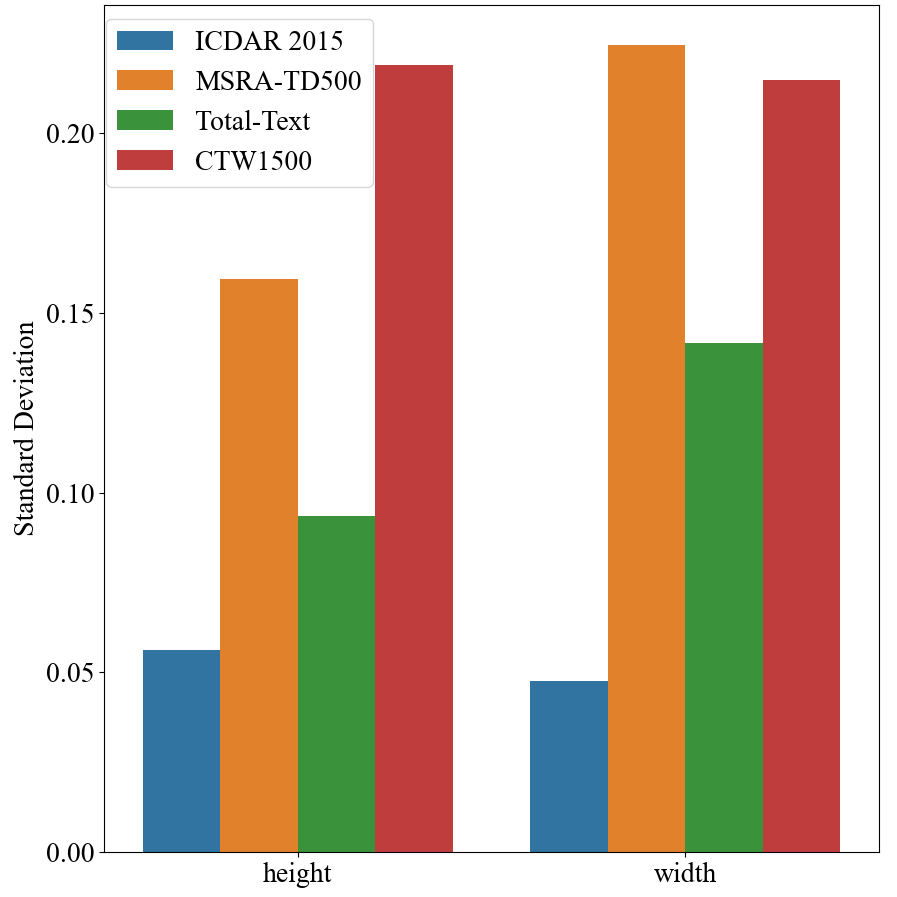}
        \caption{The standard deviation of height and width}
   \end{subfigure} %
  \qquad
   \begin{subfigure}{0.21\linewidth}
    \centering
        \includegraphics[width=\linewidth]{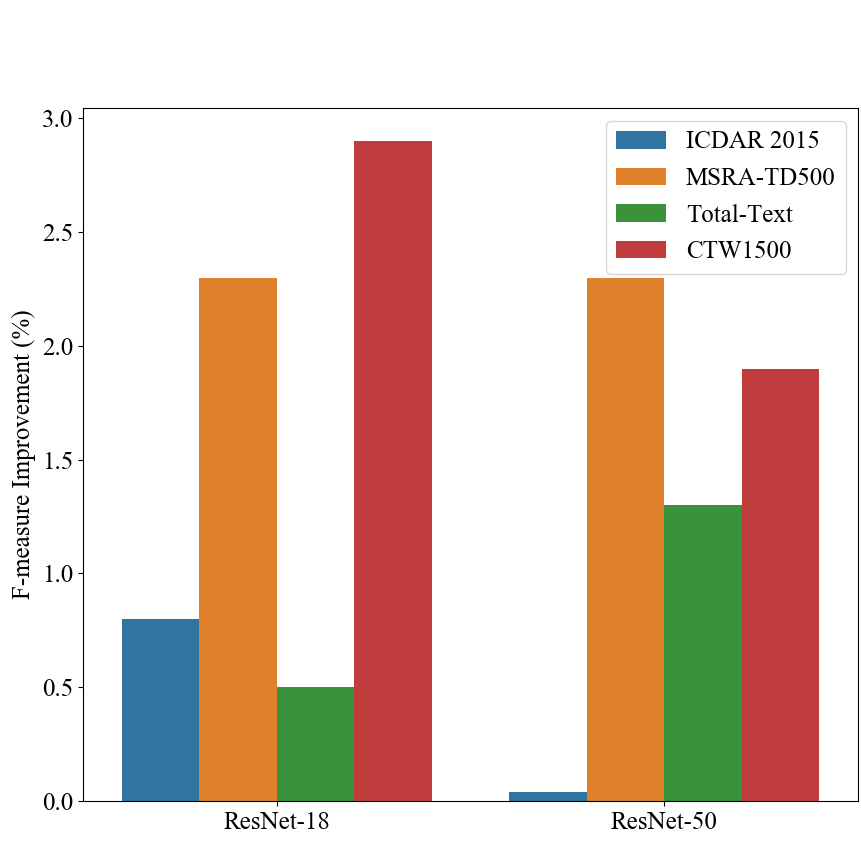}
        \caption{The improvements obtained by ASF}
   \end{subfigure}
\caption{The scale distributions of the test sets of different datasets. \revise{The points in (a) and (b) represent the text instances of various scales.} The scales of the bounding boxes are measured by the width and height of their minimum bounding rectangles.}
\label{fig:dataset_scale}
\end{figure*}

\subsection{Comparisons with the Conference Version}
The major extension of this paper over the conference version is the proposed ASF module. Some qualitative results are shown in Fig.~\ref{fig:vis_compare} \revise{and more results are shown in the appendix}. \revise{As shown, DBNet++ performs better in detecting the text instances of diverse scales, especially for the large-scale text instances. In contrast, DBNet may generate inaccurate bounding boxes or discrete bounding boxes for large-scale text instances. This indicates that the proposed ASF module strengthens the scale robustness of the text detection model.}

The quantitative results in the standard scene text benchmarks show that the proposed DBNet++ outperforms the conference version in terms of accuracy with little speed drop. Specifically, The accuracy increases 0.5\% (1.3\%), 2.9\% (1.9\%), 0.8\% (0.0\%), 3.6\% (2.3\%), and 1.3\% (1.0\%) in terms of F-measure on the Total-Text dataset, the CTW1500 dataset, the ICDAR 2015 dataset, the MSRA-TD500 dataset, and the MLT-2019 dataset, respectively, with the backbone of ResNet-18 (ResNet-50).

The performance improvements on the CTW1500 dataset and the MSRA-TD500 dataset are more significant than those on the Total-Text dataset and the ICDAR 2015 dataset. \revise{Thus, we visualize the scale distributions of these datasets in Fig.~\ref{fig:dataset_scale} for further analysis. As shown in Fig.~\ref{fig:dataset_scale}, the scale distributions of the ICDAR 2015 dataset and the Total-Text dataset are less diverse than those of the MSRA-TD500 dataset and the CTW1500 dataset. As shown in Fig.~\ref{fig:dataset_scale} (c) and Fig.~\ref{fig:dataset_scale} (d), the performance improvements approximately have a positive correlation with the diversity of the scales, which quantitatively reflects that DBNet++ is superior to DBNet on scale robustness.}

\subsection{Limitation}
One limitation of our method is that it is difficult to deal with cases ``text inside text", which means that a text instance is inside another text instance.
Although the shrunk text regions are helpful to the cases that the text instance is not in the center region of another text instance, it fails when the text instance is exactly located in the center region of another text instance. This is a common limitation for segmentation-based scene text detectors.

\section{Conclusion}\label{sec:conclusion}
In this paper, we have presented a novel framework for detecting arbitrary-shape scene text, which improves the segmentation-based scene text detection methods from two aspects: (1) A differentiable binarization module is proposed to integrate the binarization process into the training period; (2) The proposed ASF module efficiently enhances the scale robustness of the segmentation network. Both two modules significantly improve the text detection accuracy. The experiments have verified that our method (ResNet-50 backbone) consistently outperforms the state-the-the-art methods on five standard scene text benchmarks, in terms of speed and accuracy. In particular, even with a lightweight backbone (ResNet-18), our method can achieve competitive performance on all the testing datasets with real-time inference speed. 


%

\ifCLASSOPTIONcompsoc
  \section*{Acknowledgments}
\else
  \section*{Acknowledgment}
\fi

This work was supported by National Key R\&D Program of China No.2018YFB1004600 and NSFC No.61733007.

\ifCLASSOPTIONcaptionsoff
  \newpage
\fi



%



{\small
\bibliographystyle{ieee}
\bibliography{references}
}

\vfill


\clearpage
\onecolumn
\appendix

\textbf{1. Qualitative Comparisons between DBNet and DBNet++}

\begin{figure*}[htbp]
\centering
    \begin{subfigure}[b]{0.98\linewidth}
    \centering
    \includegraphics[width=0.95\textwidth]{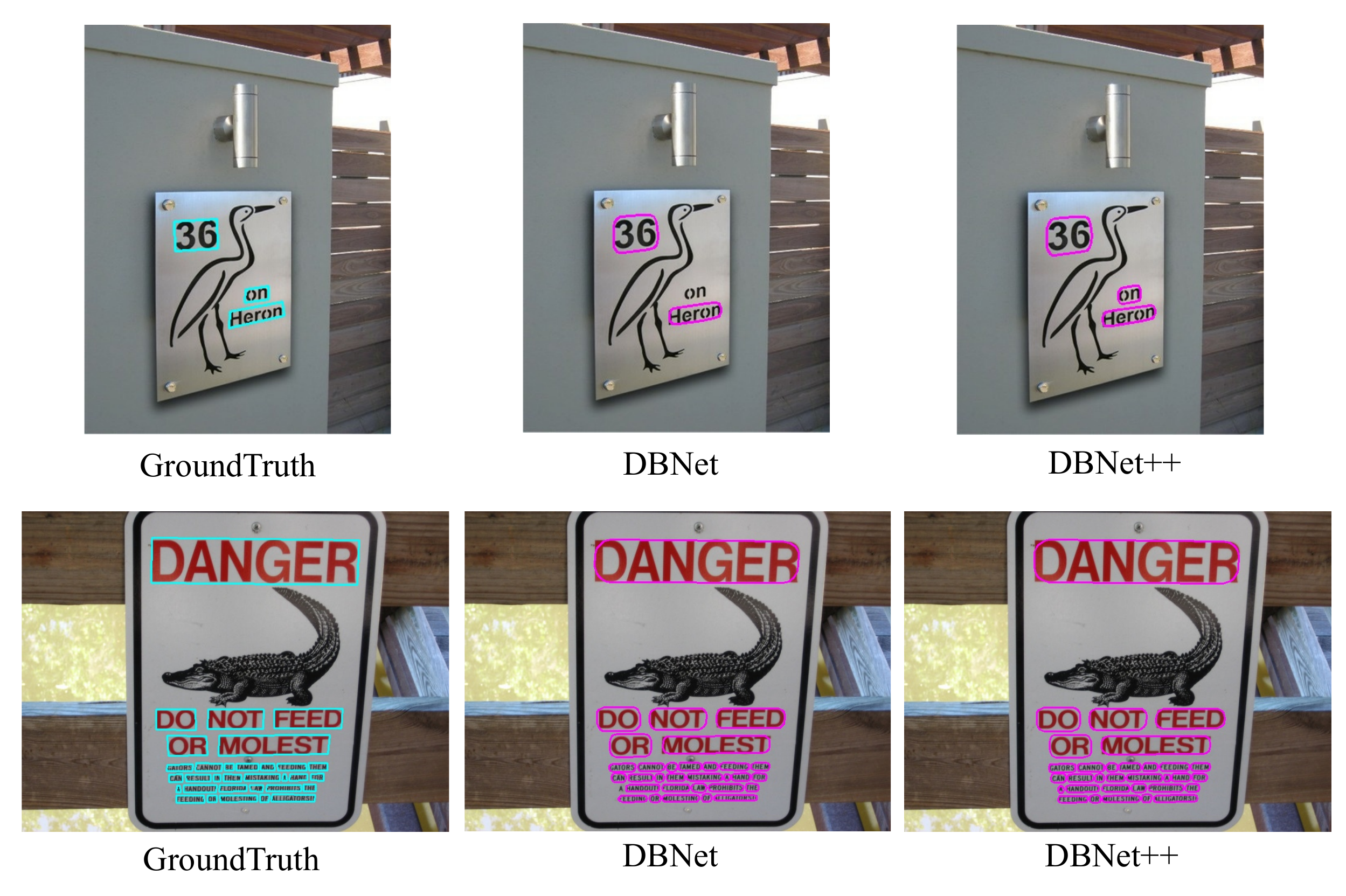}
    \label{fig:compare_1}
    \end{subfigure}
    
    \begin{subfigure}[b]{0.98\linewidth}
    \centering
    \includegraphics[width=0.95\textwidth]{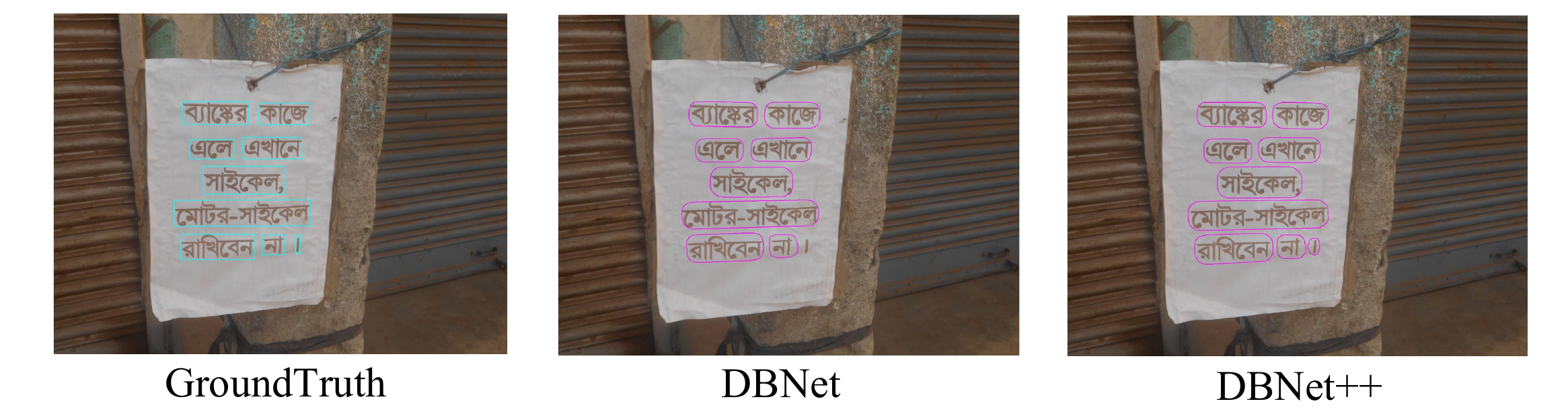}
    \label{fig:compare_5}
    \end{subfigure}
\end{figure*}    

\begin{figure*}[htbp]
\centering
    \begin{subfigure}[b]{0.98\linewidth}
    \centering
    \includegraphics[width=0.95\textwidth]{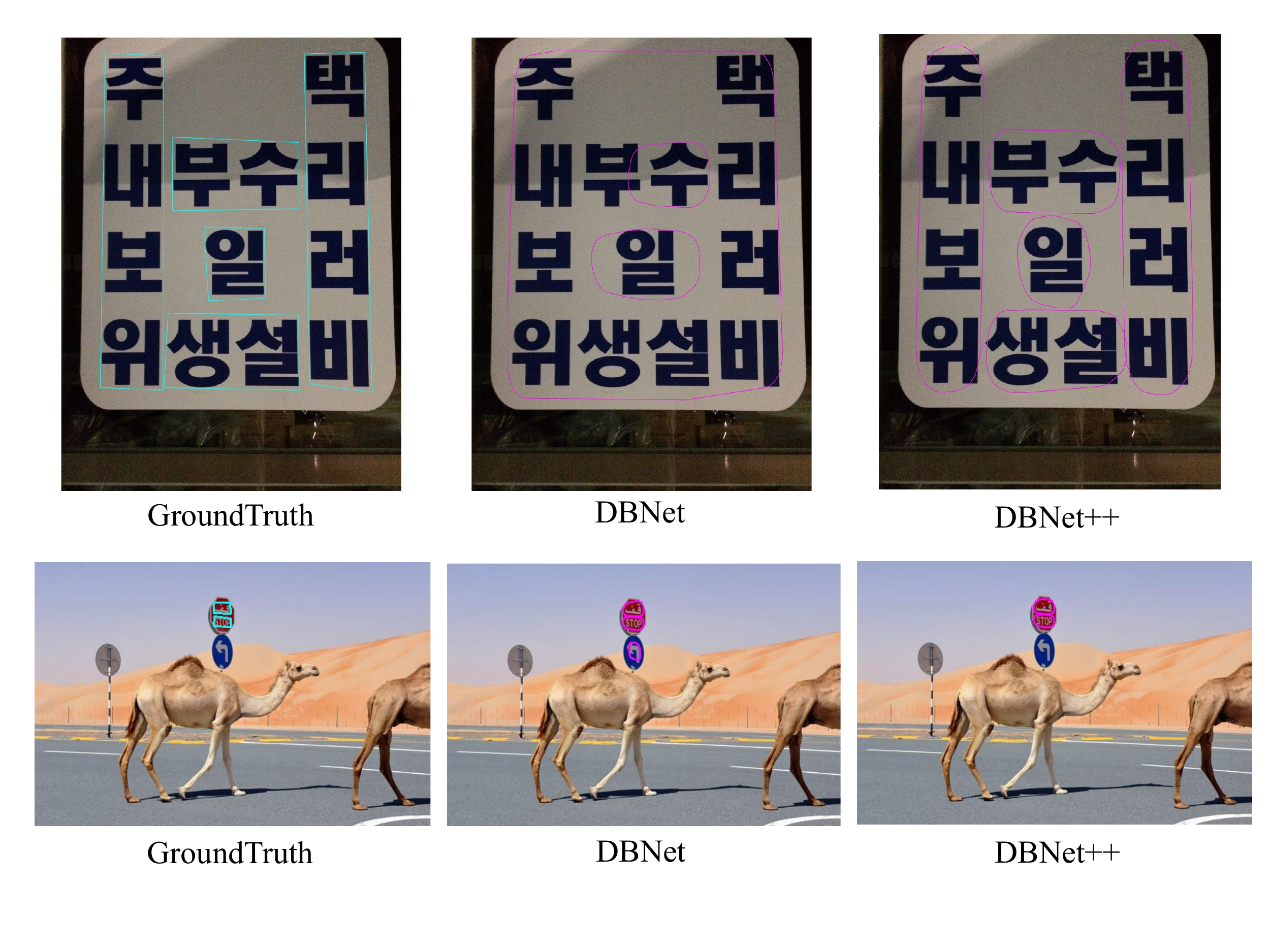}
    \label{fig:compare_2}
    \end{subfigure}
    
    \begin{subfigure}[b]{0.98\linewidth}
    \centering
    \includegraphics[width=0.95\textwidth]{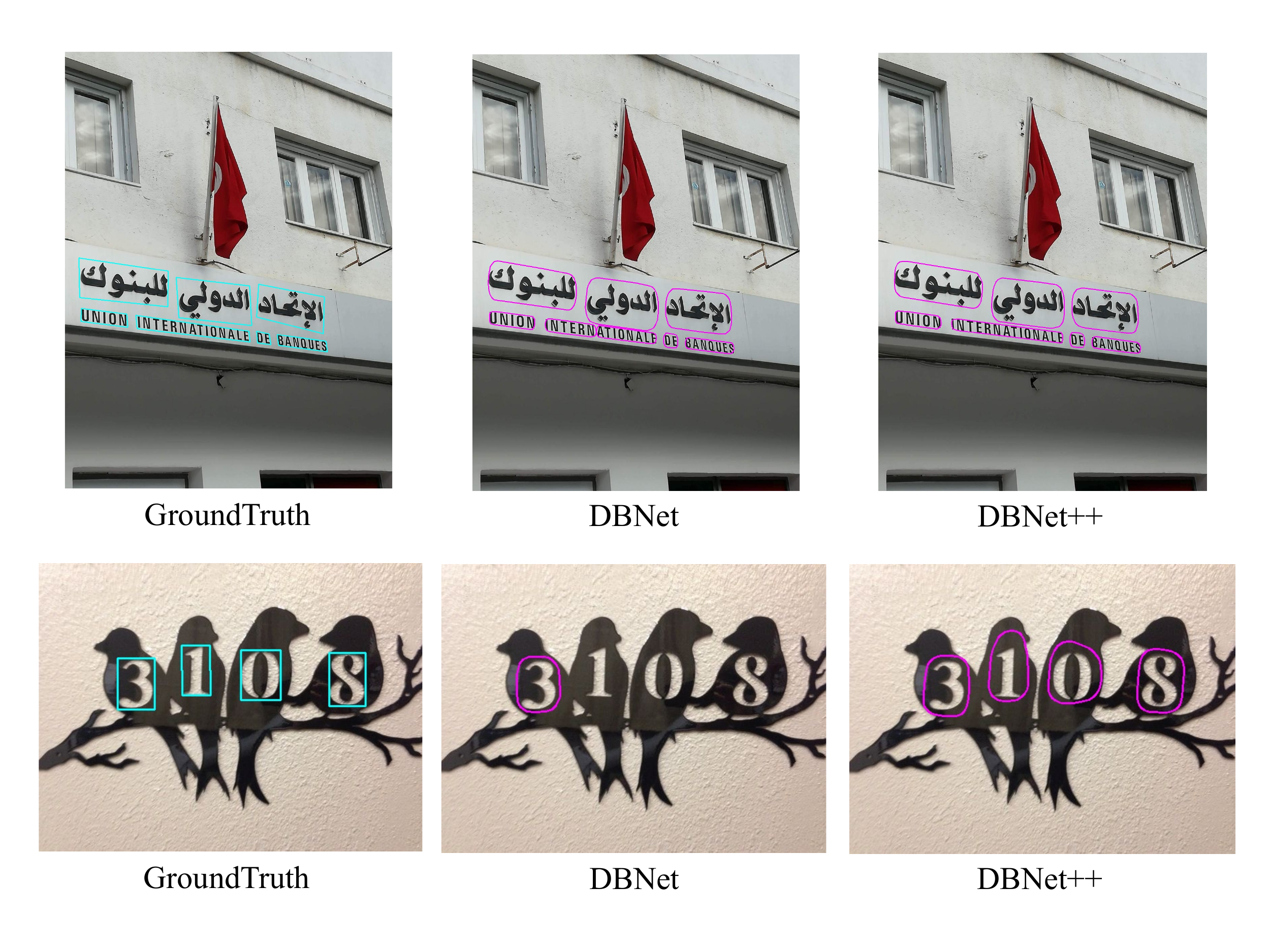}
    \label{fig:compare_3}
    \end{subfigure}
\end{figure*}

\end{document}